\documentclass{article}

\PassOptionsToPackage{square,numbers}{natbib}

\usepackage[preprint]{neurips_2025}

\usepackage[utf8]{inputenc} % allow utf-8 input
\usepackage[T1]{fontenc}    % use 8-bit T1 fonts
\usepackage{hyperref}       % hyperlinks
\usepackage{url}            % simple URL typesetting
\usepackage{booktabs}       % professional-quality tables
\usepackage{amsfonts}       % blackboard math symbols
\usepackage{nicefrac}       % compact symbols for 1/2, etc.
\usepackage{microtype}      % microtypography
\usepackage{xcolor}         % colors

\usepackage{graphicx}
\usepackage{booktabs}
\usepackage{graphicx}
\usepackage{amssymb}
\usepackage{booktabs}
\usepackage{dsfont}
\usepackage{amsfonts,amssymb}
\usepackage{graphics}
\usepackage{tikz}
\usepackage{multirow, graphicx}
\usepackage{colortbl}
\usepackage{bbding}
\usepackage{bm}
\usepackage{algorithm}
\usepackage{algorithmic}
\usepackage{amsmath}
\usepackage{multirow, graphicx}
\usepackage{colortbl}
\usepackage{bbding}
\usepackage{bm}
\usepackage{arydshln}

\usepackage{wrapfig}
\usepackage{color,xcolor, colortbl}
\definecolor{red1}{RGB}{192,0,0}
\definecolor{green1}{RGB}{58,128,37}
\usepackage{adjustbox}
\usepackage{pifont} 
\usepackage{overpic}
\usepackage{subfloat}
\usepackage{enumitem}
\usepackage{makecell}
\usepackage{array}
\usepackage{caption}
\usepackage{algorithm}
\usepackage{algorithmic}
\usepackage{textcomp}

\usepackage{multirow}
\usepackage{xspace}
\usepackage{caption}
\usepackage{thm-restate}
\usepackage{footmisc}

\newcommand{\ie}{\textit{i}.\textit{e}.}

\newcommand{\cond}{c}

\usepackage{amsmath}

\newcommand{\ours}{ScalingNoise\xspace}
\newcommand\encircle[2][]{\tikz[overlay]\node[fill=blue!20,inner sep=2pt, anchor=text, rectangle, rounded corners=1.5mm,#1] {#2};\phantom{#2}}
\definecolor{multisample}{rgb}{0.94, 0.97, 1.0}
\definecolor{video generation}{RGB}{233, 196, 107}
\definecolor{scaling}{RGB}{130, 178,154}
\definecolor{global-optimal}{RGB}{246, 111, 105}
\definecolor{efficiency}{RGB}{033, 158,188}
\definecolor{strings}{RGB}{220, 20, 60}

\title{ScalingNoise: Scaling Inference-Time Search for Generating Infinite Videos}

\author{Haolin Yang$^{1,3\ast}$, Feilong Tang$^{1,2,3\ast}$, Ming Hu$^{1,2,3}$, Qingyu Yin$^{4}$, Yulong Li$^{1,3}$, Yexin Liu$^{5}$, \\ \textbf{Zelin Peng}$^{6}$, \textbf{Peng Gao}$^3$, \textbf{Junjun He}$^3$, \textbf{Zongyuan Ge}$^{2\dag}$, \textbf{Imran Razzak}$^{1 \dag}$ \\ 
$^1$MBZUAI,
$^2$Monash University,
$^3$Shanghai AI Lab,
$^4$Zhejiang University,\\
$^5$HKUST,
$^6$Shanghai Jiaotong University,
\\ 
\normalsize Project Page: \href{https://yanghlll.github.io/ScalingNoise.github.io/}{\textit{https://yanghlll.github.io/ScalingNoise.github.io/}}
}

\begin{document}

\maketitle

\vspace{-0.5cm}
\begin{abstract}
Video diffusion models (VDMs) facilitate the generation of high-quality videos, with current research predominantly concentrated on scaling efforts during training through improvements in data quality, computational resources, and model complexity. However, inference-time scaling has received less attention, with most approaches restricting models to a single generation attempt. Recent studies have uncovered the existence of “golden noises" that can enhance video quality during generation. Building on this, we find that guiding the scaling inference-time search of VDMs to identify better noise candidates not only evaluates the quality of the frames generated in the current step but also preserves the high-level object features by referencing the anchor frame from previous multi-chunks, thereby delivering long-term value. Our analysis reveals that diffusion models inherently possess flexible adjustments of computation by varying denoising steps, and even a one-step denoising approach, when guided by a reward signal, yields significant long-term benefits. Based on the observation, we propose~\textbf{ScalingNoise}, a plug-and-play inference-time search strategy that identifies golden initial noises for the diffusion sampling process to improve global content consistency and visual diversity. Specifically, we perform one-step denoising to convert initial noises into a clip and subsequently evaluate its long-term value, leveraging a reward model anchored by previously generated content. Moreover, to preserve diversity, we sample candidates from a tilted noise distribution that up-weights promising noises. In this way, ScalingNoise significantly reduces noise-induced errors, ensuring spatiotemporal coherence in video generation. Extensive experiments on benchmark datasets demonstrate that ScalingNoise effectively improves both content fidelity and subject consistency for resource-constrained long video generation.
\end{abstract}

\vspace{-0.5cm}
\section{Introduction}
\vspace{-0.1cm}
\label{sec:introduction}
Long video generation has a significant impact on various applications, including film production, game development, and artistic creation~\cite{liu2024sora,xing2023survey,yang2024tv}. Compared to image generation~\cite{wang2023internvid,long2024videodrafter,esser2023structure}, video generation demands significantly greater data scale and computational resources due to the high-dimensional nature of video. This necessitates a trade-off between limited resources and model performance for Video Diffusion Models (VDMs)~\cite{li2024training,menapace2024snap,wu2025freeinit,tan2024videoinfinitydistributedlongvideo}.

\footnotetext[1]{Equal Contribution. $^\dagger$ Corresponding Authors.}

Recent VDMs typically adopt two main methods: one is the chunked autoregressive strategy~\cite{ho2022video,he2022latent,voleti2022mcvd,luo2023videofusion,blattmann2023align}, which predicts several frames in parallel conditioned on a few preceding ones, consequently reducing the computational burden, and “diagonal denoising” from FIFO-diffusion~\cite{kim2025fifo,chen2025ouroboros}, which re-plans the time schedule and maintains a queue with progressively increasing noise levels for denoising. However, video generation is induced by both the diffusion strategies and the noise. Variations in the noises can lead to substantial changes in the synthesized video, as even minor alterations in the noise input can dramatically influence the output~\citep{xu2024goodseedmakesgood,qi2024noisescreatedequallydiffusionnoise,zhou2024golden}. This sensitivity underscores that noises affect both the overall content and the subject consistency of video generation.

% \vspace{-0.7cm}
\begin{figure}[t]
    \centering
    \centerline{\includegraphics[width=\linewidth]{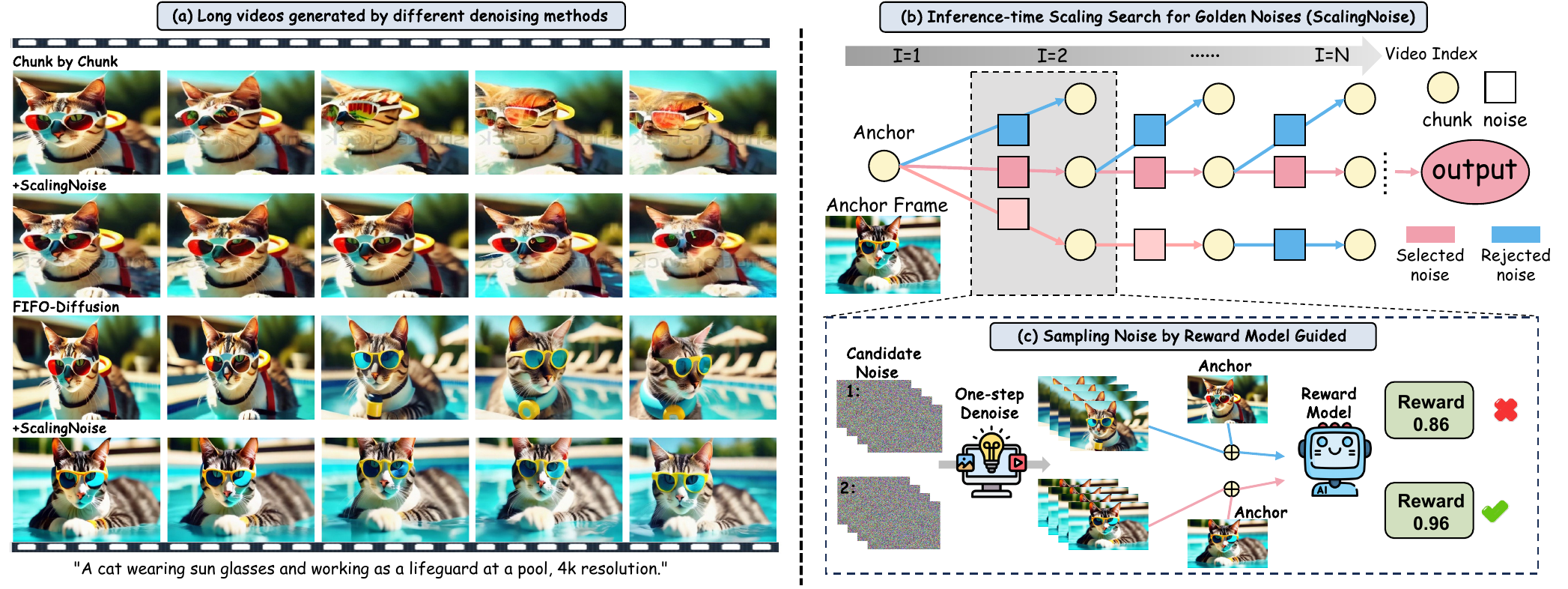}}
    \vspace{-0.1cm}
    \captionof{figure}{An overview of how \ours\ improves long video generation through inference-time search.
    (a) Chunk-by-chunk and FIFO-Diffusion methods often suffer from accumulated errors and visual degradation over long sequences.
    (b) \ours\ mitigates this by conducting a tailored step-by-step beam search for suitable initial noises, guided by a reward model that incorporates an anchor frame to ensure a long-term signal.
    (c) At each step, we perform one-step denoising on candidate noises to obtain a clearer clip for evaluation; the reward model then predicts the long-term value of each candidate, helping avoid noises that could introduce future inconsistencies.}
    \label{fig:teaser}
    \vspace{-0.4cm}
\end{figure}

The key to enhancing the quality of long video generation lies in identifying “golden noises” for the diffusion sampling process. Recent studies employ the approach of increasing data~\citep{chefer2023attendandexciteattentionbasedsemanticguidance,guo2024initnoboostingtexttoimagediffusion,xiong2024lvd2mlongtakevideodataset,yin2024precisescalinglawsvideo}, computational resources~\cite{zhao2025realtimevideogenerationpyramid,yin2025slowbidirectionalfastautoregressive,xia2025trainingfreeadaptivesparseattention,zhang2025sageattention2efficientattentionthorough,zhang2025flashvideoflowingfidelityefficient,jin2024pyramidalflowmatchingefficient}, and model size~\cite{ma2025stepvideot2vtechnicalreportpractice,nvidia2025cosmosworldfoundationmodel,lin2024opensoraplanopensourcelarge,hacohen2024ltxvideorealtimevideolatent,zhou2024allegroopenblackbox} to reduce the truncation errors during the sampling process, but these methods often incur substantial additional costs. Conversely, other approaches focus on training-free denoising strategies such as FreeNoise~\cite{zhou2024golden,qiu2024freenoise,wu2023freeinit,lu2024freelong} and Gen-L-Video~\cite{wang2023genl}. They aim to enhance the consistency of generated video by refining local denoising processes to mitigate noise-induced discrepancies, thereby ensuring smoother temporal transitions. Recently, in Large Language Models (LLMs), the study on improving their capability has expanded to inference-time~\cite{o1blog, lightman2023letsverifystepstep,yang2024qwen25mathtechnicalreportmathematical,snell2024scaling}. By allocating more compute during inference, often through sophisticated search processes, these works show that LLMs can produce higher-quality and more appropriate responses. As a result, inference-time scaling opens new avenues for improving model performance when additional resources become available after training. Similarly, recent explorations in diffusion models have leveraged the extra inference-time compute to refine noise search and enhance denoising, thereby improving sample quality and consistency~\citep{ma2025inferencetimescalingdiffusionmodels,oshima2025inference,wang2025remasking}. However, while previous works have shown effectiveness, they focus solely on information within a local window, overlooking long-term feedback and accumulated errors. In this study, we argue that scaling inference-time search of VDMs to identify golden noises enhances long-term consistencies in long video generation.

\begin{wraptable}{r}{0.5\textwidth}
    \centering
    \vspace{-0.35cm}
    \resizebox{\linewidth}{!}{
    \begin{tabular}{lcccl}
        \toprule
        \makecell{\textbf{Search}} & \makecell{\textbf{Representative Methods}}& \textbf{Type} &\textbf{Advantage}
        \\
        \midrule
        \multirow{1}{*}{\textbf{Greedy}} &Chunk-Wise Generation ~\citep{zhang2024chunkwisegenerationlongvideos} & Video & \multirow{1}{*}{\encircle[fill=efficiency, text=white]{E} \hspace{0.2 mm} }
        \\    
        \midrule 
        \multirow{3}{*}{\textbf{Tree}}&  Scaling Denoising Steps~\citep{ma2025inferencetimescalingdiffusionmodels} & Image & \multirow{2}{*}{\encircle[fill=scaling, text=white]{S} \hspace{0.2 mm} \encircle[fill=efficiency, text=white]{E}\hspace{1.2 mm}} \\
        & {Steering Generation~\cite{singhal2025generalframeworkinferencetimescaling}} & Image
        \\
        \cmidrule{2-4}
        & \ours (Ours) & Video & \encircle[fill=global-optimal, text=white]{G}\hspace{1 mm} \encircle[fill=scaling, text=white]{S} \hspace{0.2 mm} \encircle[fill=efficiency, text=white]{E}\hspace{1.2 mm} \\
         \bottomrule
    \end{tabular}
    }
    \vspace{0.1cm}
    \caption{Greedy decoding is efficient but easily trapped in local optima. Tree methods, better for global optimal decisions, is suitable for inference-time scaling. Our \ours achieves: \encircle[fill=global-optimal, text=white]{G}\ lobal-Optimality of solution,  \encircle[fill=scaling, text=white]{S}\ caling to long-range planning, and  \encircle[fill=efficiency, text=white]{E}\ fficiency.}
    \label{table: method comparison}
    \vspace{-0.3cm}
\end{wraptable}

To this end, we propose \textbf{ScalingNoise}, a plug-and-play inference-time search strategy that identifies golden initial noises by leveraging a reward model to steer the diffusion process, as illustrated in Fig.~\ref{fig:teaser} (b). Specifically, we employ beam search~\cite{xie2023selfevaluationguidedbeamsearch} tailored for intermediate steps and mitigate the accumulated error at each step, while progressively selecting the golden initial noises by choosing the initial noises. Moreover, rather than relying solely on the short-term reward of locally noised clips, we predict the long-term consequences of the initial noises to maintain high coherence. A key challenge lies in the impracticality of directly assessing the initial noises, as it typically requires multiple denoising steps to produce a clear image for assessment, resulting in an exponential increase in computational cost. 
To address this, we introduce a one-step evaluation strategy that employs the predicted clearer clip from the first DDIM step as an efficient proxy of the quality of a fully denoised clip, as illustrated in Fig.~\ref{fig:teaser} (c). Then, the predicted clip is fed into the reward model while preceding image serve as anchor, providing subject contextual information that preserves appearance consistency and supports long-term value estimation beyond the immediate search step. Our integrated search strategy achieves a practical balance between global optimality, scalability, and efficiency, as shown in Table~\ref{table: method comparison}. Moreover, while greedy decoding easily becomes trapped in local optima, the tree search strategy retains multiple candidate sequences, thus exploring the search space more comprehensively and enhancing both the quality and diversity of the generated results.

Given limited computational resources, our strategy selects from a finite pool of candidate noises. Moreover, the quality of candidate noises constitutes a critical factor, complementing the robust reward model that provides long-term supervisory signals. To ensure the candidate pool comprises noise that enhances video consistency, we construct it by sampling from a tilted distribution. Specifically, the weight of high-reward samples is increased, while samples are still drawn from the standard normal distribution to preserve diversity. Through this iterative search process, we significantly reduce accumulated errors and avoid inconsistencies arising from the randomness of initial noises.

% We validate our method's effectiveness across multiple benchmarks. Our key contributions are:
On multiple benchmarks, we verify the effectiveness of our method. Our main contributions are: \textit{(i)} We propose a plug-and-play inference-time scaling strategy for long video generation by incorporating long-term supervision signals into the reward process. \textit{(ii)} We introduce a one-step denoising approach that transforms the evaluation of initial noises into the evaluation of a clearer image without extra computational overhead. \textit{(iii)} Extensive experiments demonstrate that our proposed ScalingNoise can be effectively applied to various video diffusion models, improving the quality of generated videos.

\vspace{-0.2cm}
\section{Methodology}
\vspace{-0.2cm}

\subsection{Preliminaries: Video Denoising Approach}

\noindent \textbf{Long Video Generation.}
We introduce the formulation of VDM for generating long videos. There are two approaches to create long videos: the Chunk-by-Chunk method and the FIFO diffusion method. In the following, we provide the specific formulations for these two paradigms, respectively:
\begin{itemize} [leftmargin=0pt, labelsep=-5pt]
\item \textbf{\quad Chunk by Chunk}:
Chunk-by-chunk is a generation paradigm~\cite{wang2023genl,yuan2025brickdiffusiongeneratinglongvideos,ma2025tuningfreelongvideogeneration} that operates through a sliding window approach, using the last few frames generated in the previous chunk as the starting point for the next chunk to continuously produce content. In this paradigm, $v_i=\{v_i^f\}_{f=1}^M$ denotes a video clip of a fixed length $M$ produced by the generated model, while $v_{i,t}$ denotes $t$ noise level of the video clip. A chunk-by-chunk step can be formalized as:
\begin{equation}
    v_{i,t-1} = \Psi \left(v_{i,t}, t, \bm{\epsilon}_{\theta}(v_{i,t},t,\cond)\right),
\end{equation}
where $\Psi$ and $\bm{\epsilon}_\theta$ denote a sampler such as DDIM and a noise predict network, respectively, and $\cond$ can be denoted as a single prompt or a prompt, image pair.
\item \textbf{\quad FIFO-Diffusion}:
Different from the aforementioned process, FIFO-Diffusion~\cite{kim2025fifo} introduces a diagonal denoising paradigm~\cite{ruhe2024rolling,NEURIPS2024_2aee1c41} by rescheduling noise steps. It achieves autoregressive generation by maintaining a queue where the noise level increases step by step. We define the queue as $Q = \left\{ v_{i,t}\right\} _{t=1}^T$, where $t$ denotes the noise level, and $i$ indicates the $i$-th frame in the queue. In this paradigm,  $i$ is equal to $t$, and $v_i$ denotes a frame. The length of $T$ is $M \times P$ where $P$ denotes the partition of the queue. The procedure of FIFO step can be described as follows:
\begin{equation}
     Q = \Psi \left(Q, \left\{ {t} \right\}_{t=1}^T, \bm{\epsilon}_{\theta}(Q,\left\{ {t} \right\}_{t=1}^T,\cond)\right).
\end{equation}
\end{itemize}

\noindent \textbf{DDIM.}
DDIM~\cite{song2020ddim} introduces a new sampling paradigm by de-Markovizing the process, which remaps the time steps of DDPM~\cite{ddpm_begin} from \([0; \dots ;T]\) to \([\tau_0; \dots ;\tau_T]\), which is a subset of the initial diffusion scheduler, thereby accelerating the sampling process of DDPM. Here, DDIM\((v_{\tau_t})\) consists of three distinct components, which can be formulated as:
\begin{equation}
        v_{\tau_{t-1}} = \text{DDIM}(v_{\tau_t}) = \alpha_{\tau_{t-1}} \left(\frac{v_{\tau_t} - \sigma_{\tau_t}\bm{\epsilon}_\theta(v_{\tau_t}, \tau_t)}{\alpha_{\tau_t}} \right) + \sigma_{\tau_{t-1}} \bm{\epsilon}_\theta(v_{\tau_t}, \tau_t),
    \label{equ:ddim_reverse}
\end{equation}
where $\alpha$ and $\sigma_{\tau_t}$ denote predefined schedules.

\subsection{Formulation of Video Generation Inference}
Long video generation paradigms can be seamlessly integrated into the subsequent framework. Consider extending a pre-trained model that generates fixed-length videos into a long video generation model with a distribution of $p_\theta$. This model processes an input to generate a video $\mathbf{v}=[v_1,v_2,\dots,v_N]$, where $\mathbf{v}$ consists of N step-level responses. Each step-level response $v_i$ is a video clip of the long video, treated as a sample drawn from a conditional probability distribution:  
\begin{equation}
    v_i = p_\theta(v_i| \mathbf{v}_{<i}, \cond),  \quad i=1,2, \dots, T,
\end{equation}
where $\mathbf{v}_{<i}=[v_1, v_2, \dots, v_i]$ denotes the concatenated video. 
Moreover, this framework can be formulated as a Markov Decision Process (MDP) problem defined by the tuple $(\mathcal{S}, \mathcal{A}, \mathcal{R})$.
$\mathcal{S}$ is the state space. Each state is defined as a combination of the generated video and the condition. The initial state $\bm{s}_0$ only corresponds to the input. $\bm{s}_i$ is the combination of the currently generated videos.
$\mathcal{A}$  denotes the action space, where each action encompasses a two-part process: sampling an initial noise from a tilted distribution, followed by denoising the current video clip based on the noise.
We also have the reward function $\mathcal{R}$ to evaluate the reward of each action, which is also known as the process reward model (PRM) in LLMs. With this MDP modeling, we can search for additional states by increasing the inference-time compute, thereby obtaining a better VDM response $\mathbf{v}$. Specifically, we can take different actions in each state, continuously explore, and then make choices based on the reward model to achieve a better state. The general formulation of the selection process is:
\begin{equation}
    \bm{a}_{t+1} = \arg\max_{\mathcal{A}} \Phi (\bm{s}_t, \mathcal{A}),
\end{equation}
where $\Phi$ denotes the reward model $\mathcal{R}: \mathcal{S} \times \mathcal{A} \to \mathbb{R}$.
The core of our method focuses on efficiently and accurately estimating then selecting initial noises, improving video generation with better guidance.

% \begin{algorithm}[t]
% \caption{\ours Inference-time Search}
% \label{alg: ScalingNoise}
% \begin{algorithmic}[1]
% \REQUIRE{
% Diffusion Model $D$, Reward Function $\Phi$, Sample Tilted Distribution $Sample$, Condition $\cond$, Beam Size $k$, Step Size $n$, DDIM Steps $\tau_t$, Generated Video $\bm{V}=[\ ]$} Anchor Frame $\bm{v}_a$
% \WHILE{Generation is not Done}
% \FOR{i in [1, 2, ..., k]}
%     \STATE $\bm{r} = [\ ]$
%     \FOR{j in [1, 2, ..., n]}
%         \STATE{$\bm{\epsilon}_{ij} \gets Sample(\bm{V}$)}
%         \STATE{$\hat{\bm{v}}_{ij} \gets D(\bm{\epsilon}_{ij}, \cond,\text{num\_steps}=\tau_t)$}
%         \STATE{$r_{ij} \gets \Phi(\hat{\bm{v}}_{ij}, \bm{v}_a)$}
%         \STATE{$\bm{r}$.append($r_{ij}$)}
%         \ENDFOR
%     \ENDFOR
% \STATE{$[\bm{v}_1, \dots, \bm{v}_k] \gets \text{Select the best } k \text{ elements from } \bm{r}$}
% \FOR{i in [$\tau_0$, $\tau_1$, ..., $\tau_t$]}
%     \STATE{$\bm{v} \gets D(\bm{\epsilon}, \cond, \text{num\_steps}=i)$}
%     \ENDFOR
% \STATE {$\bm{v}_a \gets \bm{v}_{i0}$}
% \STATE Append current clip $[\bm{v}_1, \dots, \bm{v}_k]$ to $\bm{V}$
% \ENDWHILE
% \RETURN{$\bm{V}$}
% \end{algorithmic}
% \end{algorithm}
% \vspace{-0.5cm}

% \vspace{3mm}
\subsection{Reward Design}
During the search process, our objective is to evaluate the consistency and quality of the video at each step, using these insights to guide subsequent searches, as illustrated in Fig.~\ref{fig:teaser} (c). The evaluation of each search step, defined as applying an action $a_t$ to the state $\bm{s}_t$, is performed by a reward function $r_t = \Phi(s_t, a_t) \in \mathbb{R}$. Below, we elaborate on the specific design of this reward function.

\noindent \textbf{One-Step Denoising.}
Throughout our evaluation process, the actions in the MDP sampling initial noises, which is typically Gaussian, are difficult to assess. To address this, we propose a one-step evaluation approach. Specifically, we utilize the $\text{predicted } {\hat{v}_{\tau_0}}$ component from Eq.~\ref{equ:ddim_reverse} above as the target for evaluation. The detailed formulation is presented as: 
\begin{equation}\small
    { \hat{v}}_{\tau_0} = \frac{v_{\tau_t} - \sigma_{\tau_t}\bm{\epsilon}_\theta(v_{\tau_t}, \tau_t)}{\alpha_{\tau_t}}.
\end{equation}
Our method uses a single DDIM step to efficiently evaluate initial noise, unlike the resource-heavy brute-force approach that fully denoises it into clear video. While less intensive, our technique may produce suboptimal results and is more practical for scaling to long video generation.

\noindent \textbf{Consistency Reward.}
\label{sec: 3.3}
After obtaining an evaluable object, we need to design a long-term reward to evaluate the overall consistency and prevent the accumulation of errors. The reward function requires careful design. Specifically, we take into account the video clip currently being generated by the action and the states of previous nodes. To this end, we select fully denoised video frames $v_a$ from several frames prior as anchor points and assess the consistency within the current window after one-step denoising. This approach substantially reduces inconsistencies in video generation. In our experiments, we employ a DINO~\cite{Caron_2021_ICCV} model and calculate the reward using the following formula:
\begin{equation}\small
    \Phi := \frac{1}{2(T-1)} \sum_{i=1}^M \left( \langle d_{\text{a}} \cdot d_i \rangle + \langle d_i \cdot d_{i-1} \rangle \right),
    \label{eq7}
\end{equation}
where $d_a$ and $d_i$ denote the image features of the anchor frame and the $i$-th frame in the current clip, respectively. And $\langle \cdot \rangle$ is the dot product operation for calculating cosine similarity.

% \vspace{3mm}
\subsection{Action Design}
\noindent \textbf{Search Framework.}
Once equipped with a rewards model (Section~\ref{sec: 3.3}), VDM can leverage any planning algorithm for search, as demonstrated in ~\cite{hao2023reasoninglanguagemodelplanning}. We employ beam search (Fig.~\ref{fig:teaser} (b)), a robust planning technique that efficiently navigates the search tree space, effectively balancing exploration and exploitation to identify high-reward trajectories. Specifically, each node represents a state and each edge denoting an action and the resulting state transition. To steer VDM toward the most promising nodes, the algorithm relies on a state-action reward function $\Phi$ in Eq.~\ref{eq7}. To promote at each step, we maintain $K$ distinct trajectories. From a tilted distribution, we sample $N$ initial noise instances, generating $K \times N$ candidates for the current step. The reward model evaluates each candidate, and the top-$K$ candidates with the highest scores are selected as responses for that step. This sampling and selection process iterates until the full response sequence is generated. Further details and the pseudo-code for this planning algorithm are provided in Algorithm~\ref{alg: ScalingNoise} in Appendix~\ref{appendix:A}.

\noindent \textbf{Tilted Distribution Sampling.}
During the search process, we need to sample the initial noises. Due to computational constraints, exhaustively searching the entire noise space is infeasible. To increase the likelihood of generating superior results, we sample initial noise from a tilted distribution~\cite{singhal2025generalframeworkinferencetimescaling,qi2024mutualreasoningmakessmaller}. Specifically, we construct a high-quality candidate pool by employing four distinct tilted distributions to sample the initial noises. These operations are detailed as follows: 

  	$\diamond$ \textit{\textbf{A1} Random Sampling: Directly sample noise from a Gaussian distribution $v_i\sim \mathcal{N}(0, I)$}.

  		$\diamond$  \textit{\textbf{A2} FFT Sampling: Utilize 2D/3D Fast Fourier Transform (FFT)~\cite{lu2024freelong,chen2025ouroboros} to blend Gaussian noise with the last few frames, denoted as:
        \begin{equation}
	v_{i} = \boldsymbol{F}_\text{low}^{r}(v_{i})+\boldsymbol{F}_\text{high}^{r}(\eta)~,
\end{equation}
where $\boldsymbol{F}$ denotes the FFT function, and $\eta$ is the Gaussian noise.}
  	
  	$\diamond$  \textit{\textbf{A3} DDIM Inversion: Apply DDIM Inversion to re-noise the previous frame formulated as:
    \begin{equation}
    v_{t-1} = \alpha_{t-1} \left(\frac{v_t - \sigma_t\epsilon_\theta(v_t, t)}{\alpha_t} \right) + \sigma_{t-1} \epsilon_\theta(v_t, t).
\end{equation}}

  	$\diamond$  \textit{\textbf{A4} Inversion Resampling: Building on DDIM Inversion, sample new noise within its $\delta$-neighborhood defined as $\{v':d(v,v')<\delta\}$.}
Through these strategies, we enhance the quality of candidate noise, enabling us, within limited resources, to maximize the leveraging of actions with higher rewards.

\vspace{-0.2cm}
\section{Experiment}
\vspace{-0.2cm}

In this section, we conduct experiments to answer the two questions: \textit{1. Does the long-term reward guided search yield higher-quality video compared with other inference-time scaling methods?} \textit{2. Does the one-step evaluation provide an efficient and accurate assessment of initial noises?}

\vspace{-0.3cm}
\subsection{Baseline and Implementation details}
\label{subsec:implementation}
\vspace{-0.3cm}

% \begin{wrapfigure}{r}{0.5\textwidth}
%     \centering
%     \includegraphics[width=0.5\textwidth]{figures/fvd_is.pdf}
%     \caption{Comparisons of $\text{FVD}_{128}$ and IS scores on UCF-101. \ours utilizes Latte~\cite{ma2024latte} as its baseline, where the number of beam sizes is 2, and noise candidates are 5. The FVD and IS scores of the other algorithms are obtained from their respective papers, and PVDM~\cite{yu2023pvdm} denotes PVDM-L (400-400s).}
%     \label{fig:fvd_is}
% \end{wrapfigure}
%     \centering
%     \includegraphics[width=1\linewidth]{figures/fvd_is.pdf}
%     \caption{Comparisons of $\text{FVD}_{128}$ and IS scores on UCF-101. \ours utilizes Latte~\cite{ma2024latte} as its baseline, where the number of beam sizes is 2, and noise candidates are 5. The FVD and IS scores of the other algorithms are obtained from their respective papers, and PVDM~\cite{yu2023pvdm} denotes PVDM-L (400-400s).}
%     \label{fig:fvd_is}
% \end{figure}
To evaluate the effectiveness and generalization capacity of our proposed method, we implement it on the text-to-video FIFO-Diffusion~\cite{kim2025fifo} and image-to-video chunk by chunk long video generation, based on existing open-source diffusion models trained on short video clips. These models are limited to producing videos with a fixed length of 16 frames. In our experiments, the evaluations are conducted on an NVIDIA A100 GPU.

% \vspace{3mm}

\noindent \textbf{Vbench Dataset.} Our approach is systematically evaluated using VBench~\cite{Huang_2024_CVPR}, a video generation benchmark featuring 16 metrics crafted to thoroughly evaluate motion quality and semantic consistency. We select 40 representative prompts spanning all categories, generate 100 video frames, and analyze the model's performance using five metrics for performance comparison: Subject Consistency, Background Consistency, Motion Smoothness, Temporal Flickering, and Imaging Quality.
We maintain k distinct beam candidates and sample n completions for each beam. Specifically, we set beam size k = 2, 3, and n = 5 with FIFO-Diffusion based on VideoCraft2~\cite{chen2024videocrafter2}, n = 10 with ConsistI2V~\cite{ren2024consisti2v} chunk by chunk to balance the quality and efficiency. For the I2V model, we use the first frame generated by FIFO-Diffusion to guide the video generation. We consider two types of baselines: (1) \textbf{Base Model}: This employs a naive method that avoids any form of inference-time scaling. (2) \textbf{Best of N (BoN)}: A widely adopted technique to improve model response quality during inference. Specifically, we generate 3 and 5 distinct outputs. We also select three state-of-the-art methods as baselines, namely StreamingT2V~\cite{henschel2024streamingt2v}, Openasora v1.1~\cite{yuan2024opensoraplan}, and FreeNoise~\cite{qiu2024freenoise}.

\begin{table*}[!t]
\centering
\resizebox{\linewidth}{!}{
\begin{tabular}{lccccccc}
\toprule
\textbf{Method}     & $\bm{N_{can}}$       & \textbf{\makecell{Subjection\\ Consistency$\uparrow$}} & \textbf{\makecell{Background\\ Consistency$\uparrow$}} & \textbf{\makecell{Motion\\ Smoothing$\uparrow$}} & \textbf{\makecell{Time\\ Flicking$\uparrow$}} & \textbf{\makecell{Imaging\\ Quality$\uparrow$}} & \textbf{\makecell{Overall\\ Score$\uparrow$}}\\

\midrule
Streamingt2v~\cite{henschel2024streamingt2v}      &    1     &    84.03                              &         91.01                          &             96.58               &             95.47             &            61.64       &     85.74    \\
OpenSora v1.1~\cite{yuan2024opensoraplan}      &     1     &     86.92                             &        93.18                           &                  97.50          &         \textbf{98.72}                 &       53.07          &  85.87            \\ 
FreeNoise~\cite{qiu2024freenoise}         &      1     &       \underline{92.30}                           &        \textbf{95.16}                           &            96.32                &           94.94               &          \underline{67.14}    & \underline{89.17}               \\ \midrule
ConsistI2V(Chunk-wise)~\cite{ren2024consisti2v}   &   1    &          89.57 \textcolor{gray}{$\downarrow$ +0.00}                       &         93.22   \textcolor{gray}{$\downarrow$ +0.00}                        &         97.62    \textcolor{gray}{$\downarrow$ +0.00}                &          96.63   \textcolor{gray}{$\uparrow$ +0.00}             &           55.21   \textcolor{gray}{$\downarrow$ +0.00}      &       86.45   \textcolor{gray}{$\downarrow$ +0.00}      \\ \hdashline
+BoN      &   3     &           89.92 \textcolor{green1}{$\uparrow$ +0.35}                   &          93.64  \textcolor{green1}{$\uparrow$ +0.42}                       &          97.59   \textcolor{red1}{$\downarrow$ $-$0.03}                &           96.66  \textcolor{green1}{$\uparrow$ +0.03}             &         55.74   \textcolor{green1}{$\uparrow$ +0.50}       & 86.71   \textcolor{green1}{$\uparrow$ +0.26}       \\
+BoN      &    5    &                   90.56  \textcolor{green1}{$\uparrow$ +0.99}             &               93.59   \textcolor{green1}{$\uparrow$ +0.37}                 &         97.73   \textcolor{green1}{$\uparrow$ +0.11}                &              96.24   \textcolor{red1}{$\downarrow$ $-$0.39}         &             56.24    \textcolor{green1}{$\uparrow$ +1.03}     &      86.87  \textcolor{green1}{$\uparrow$ +0.42}   \\ \rowcolor{gray!30}
+\ours (Ours)   &   2        &     91.58     \textcolor{green1}{$\uparrow$ +2.01}                        &           94.36     \textcolor{green1}{$\uparrow$ +1.14}                   &             \underline{97.85}     \textcolor{green1}{$\uparrow$ +0.25}          &            96.79    \textcolor{green1}{$\uparrow$ +0.16}          &         56.82  \textcolor{green1}{$\uparrow$ +1.61}       & 87.48   \textcolor{green1}{$\uparrow$ +1.03}        \\ \rowcolor{gray!30}
+\ours (Ours)  &    3        &             92.02    \textcolor{green1}{$\uparrow$ +2.45}                 &           94.44   \textcolor{green1}{$\uparrow$ +1.22}                     &             \textbf{97.91}  \textcolor{green1}{$\uparrow$ +0.29}             &         \underline{96.97}    \textcolor{green1}{$\uparrow$ +0.34}             &                   58.12  \textcolor{green1}{$\uparrow$ +2.91}    &    87.89 \textcolor{green1}{$\uparrow$ +1.18} \\ \midrule
FIFO-Diffusion~\cite{kim2025fifo}   &    1            &         90.26   \textcolor{gray}{$\downarrow$ +0.00}                       &        93.53       \textcolor{gray}{$\downarrow$ +0.00}                     &        95.86    \textcolor{gray}{$\downarrow$ +0.00}                 &     92.78             \textcolor{gray}{$\downarrow$ +0.00}         &           65.52  \textcolor{gray}{$\downarrow$ +0.00}     &      87.59   \textcolor{gray}{$\downarrow$ +0.00}    \\ \hdashline
+BoN   &    3               &        90.92  \textcolor{green1}{$\uparrow$ +0.66}                        &        94.49    \textcolor{green1}{$\uparrow$ +0.96}                       &          95.20   \textcolor{red1}{$\downarrow$ $-$0.66}               &           93.76   \textcolor{green1}{$\uparrow$ +0.98}            &         64.13  \textcolor{red1}{$\downarrow$ $-$1.39}       &  87.70  \textcolor{green1}{$\uparrow$ +0.11}         \\
+BoN    &    5               &      91.26   \textcolor{green1}{$\uparrow$ +1.00}                         &            94.91    \textcolor{green1}{$\uparrow$ +1.38}                   &          95.97 \textcolor{green1}{$\uparrow$ +0.11}                 &                93.97   \textcolor{green1}{$\uparrow$ +1.19}       &           64.37  \textcolor{red1}{$\downarrow$ $-$1.15}     &  88.10 \textcolor{green1}{$\uparrow$ +0.51}          \\ \rowcolor{gray!30}
+\ours (Ours)    &    2                 &      91.60     \textcolor{green1}{$\uparrow$ +1.34}                       &         93.74    \textcolor{green1}{$\uparrow$ +0.21}                      &                      96.67  \textcolor{green1}{$\uparrow$ +0.81}    &           94.96     \textcolor{green1}{$\uparrow$ +2.18}          &          65.67  \textcolor{green1}{$\uparrow$ +0.15}      &  88.53 \textcolor{green1}{$\uparrow$ +0.94}          \\
\rowcolor{gray!30}
+\ours (Ours)  &    3             &      \textbf{93.14}   \textcolor{green1}{$\uparrow$ +2.88}                         &         \underline{94.61}   \textcolor{green1}{$\uparrow$ +1.08}                       &                       97.01  \textcolor{green1}{$\uparrow$ +1.15}   &           95.34   \textcolor{green1}{$\uparrow$ +2.56}            &            \textbf{67.91} \textcolor{green1}{$\uparrow$ +2.39}       &  \textbf{89.60}   \textcolor{green1}{$\uparrow$ +2.01}       \\ 

\bottomrule

\end{tabular}
}
\vspace{-0.2cm}
\caption{\textbf{Quantitative comparison results.} Comparison of performance metrics for various video generation methods as benchmarked by VBench. We calculate the average performance in the last column, demonstrating its effectiveness in producing fidelity and consistent long videos. Bold indicates the highest value, and underlined indicates the second highest.}
\vspace{-2mm}
\label{tab:mainexp}
\end{table*}

% \vspace{3mm}
\noindent \textbf{UCF-101 Dataset.}
UCF-101~\cite{soomro2012ucf101} is a large-scale human action dataset containing 13,320 videos across 101 action classes. We utilize the following metrics: 
\begin{itemize}
[leftmargin=0pt, labelsep=-5pt]
    \item \quad Frechet Video Distance (FVD)~\cite{unterthiner2019fvd} for temporal coherence and motion realism.
    \item \quad Inception Score (IS)~\cite{salimans2016improved} for frame-level quality and diversity.
\end{itemize}
We measure the two metrics using Latte~\cite{ma2024latte} as a base model, which is a DiT-based video model trained on UCF-101~\cite{soomro2012ucf101}, employing FIFO-Diffusion as the paradigm for long video generation, configured with k = 2 and m = 5. We generate 2,048 videos with  128 frames each to calculate $\text{FVD}_{128}$, a specialized
version of FVD which uses 128-frames-long videos to compute the statistics, and randomly sample a 16-frame clip from each video to measure the IS score, following evaluation guidelines in StyleGAN-V~\cite{skorokhodov2022styleganv}. As the base model, we choose StyleGAN-V, PVDM-L (400-400s)~\cite{yu2023pvdm}, FIFO-Diffusion, as they are three representative open-sourced models.  

\vspace{-0.2cm}
\subsection{Quantitative results}
\vspace{-0.2cm}
\begin{wrapfigure}{r}{0.5\textwidth}
    \vspace{-4mm}
    \centering
    \includegraphics[width=0.5\textwidth]{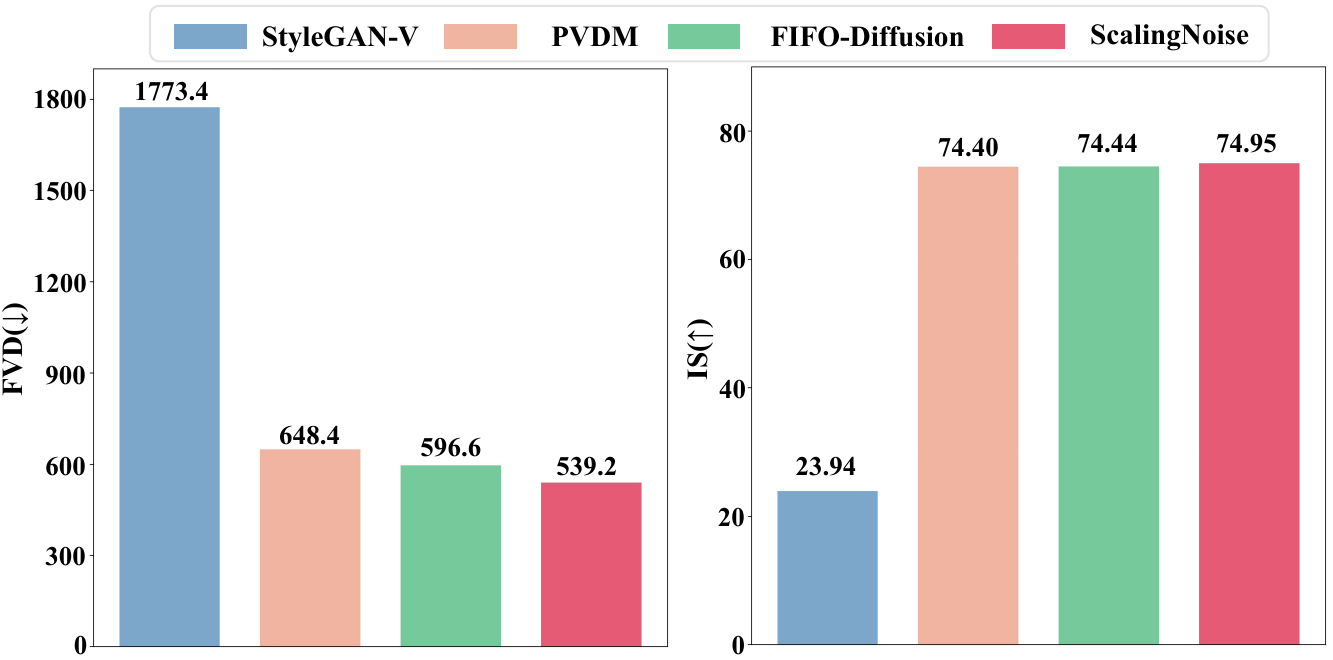}
    \caption{Comparisons of $\text{FVD}_{128}$ and IS scores on UCF-101. \ours utilizes Latte~\cite{ma2024latte} as its baseline, where the number of beam sizes is 2, and noise candidates are 5. The FVD and IS scores of the other algorithms are obtained from their respective papers, and PVDM~\cite{yu2023pvdm} denotes PVDM-L (400-400s).}
    \label{fig:fvd_is}
    \vspace{-2mm}
\end{wrapfigure}
\textbf{Scale Search Improves Video Consistency.}\label{sec: 4.1.1} We compare \ours with the baselines in terms of multiple benchmarks. \textbf{Vbench}: As shown in Table~\ref{tab:mainexp}, we find that the videos generated by \ours are significantly more preferred compared with the baseline. While increasing inference compute via BoN shows improvement, they still fall short compared with \ours. Although its consistency has improved, our method outperforms BoN(5) across all evaluated aspects. The long videos obtained using \ours search significantly \emph{ehnance consistency} and provide \emph{high quality} video frames.  \textbf{FVD and IS}: As illustrated in Fig.~\ref{fig:fvd_is}, our approach outperforms all the compared methods including PVDM-L (400-400s)~\cite{yu2023pvdm}, which employs a chunked autoregressive generation strategy. Note that PVDM-L iteratively generates 16 frames conditioned on the previous outputs over 400 diffusion steps.

% \vspace{3mm}
\noindent \textbf{Benefits from Further Scaling Up Inference Compute.} We next explore the effect of increasing inference-time computation on the response quality at each step by varying the beam sizes. For fairness, we set n = 5 for FIFO-Diffusion and n = 10 for chunk-by-chunk methods, reflecting their differing noise initialization needs.
This difference results in a search space complexity significantly larger than that of FIFO-Diffusion. We report the scores for long video generation achieved through \ours search, based on both paradigms, with beam sizes set from 1 to 4. The experimental results are illustrated in Fig.~\ref{fig:largefig} (a). Since some prompts are static while others correspond to video actions with very large movements, this results in a significant variance. Our observations reveal that the performance of \ours, for both strategies, improves steadily as the search beam size increases. This trend suggests that scaling inference-time computation effectively enhances the visual consistency capabilities of VDM.

% \begin{figure}[!t]
%   \centering
%   \includegraphics[width=\linewidth]{figures/exp.pdf}
%   \vspace{-0.5cm}
%   % \caption{The two images on the left are boxplots showing the effect of increasing scale, with the leftmost based on a FIFO implementation of scale noise and the adjacent one using a chunk-by-chunk approach. The two images on the right depict search results guided by different reward models, with the left one guided by DINO and the right one guided by CLIP.}
%   \caption{(a) The two figures are boxplots showing the tendency of scaling beam sizes for ScaleNoise based on two paradigms, in order of FIFO-Diffusion and Chunk by chunk. (b) From Left to Right: Correction of reward model DINO and CLIP feature similarity score and final subject consistency. All points are generated by VideoCraft2.}
%     \label{fig:largefig}
%   \vspace{-0.2cm}
% \end{figure}
\begin{figure}[!t]
    \centering
    \begin{minipage}{0.48\textwidth}
        % \vspace{-2cm}
        \centering
        \includegraphics[width=\textwidth]{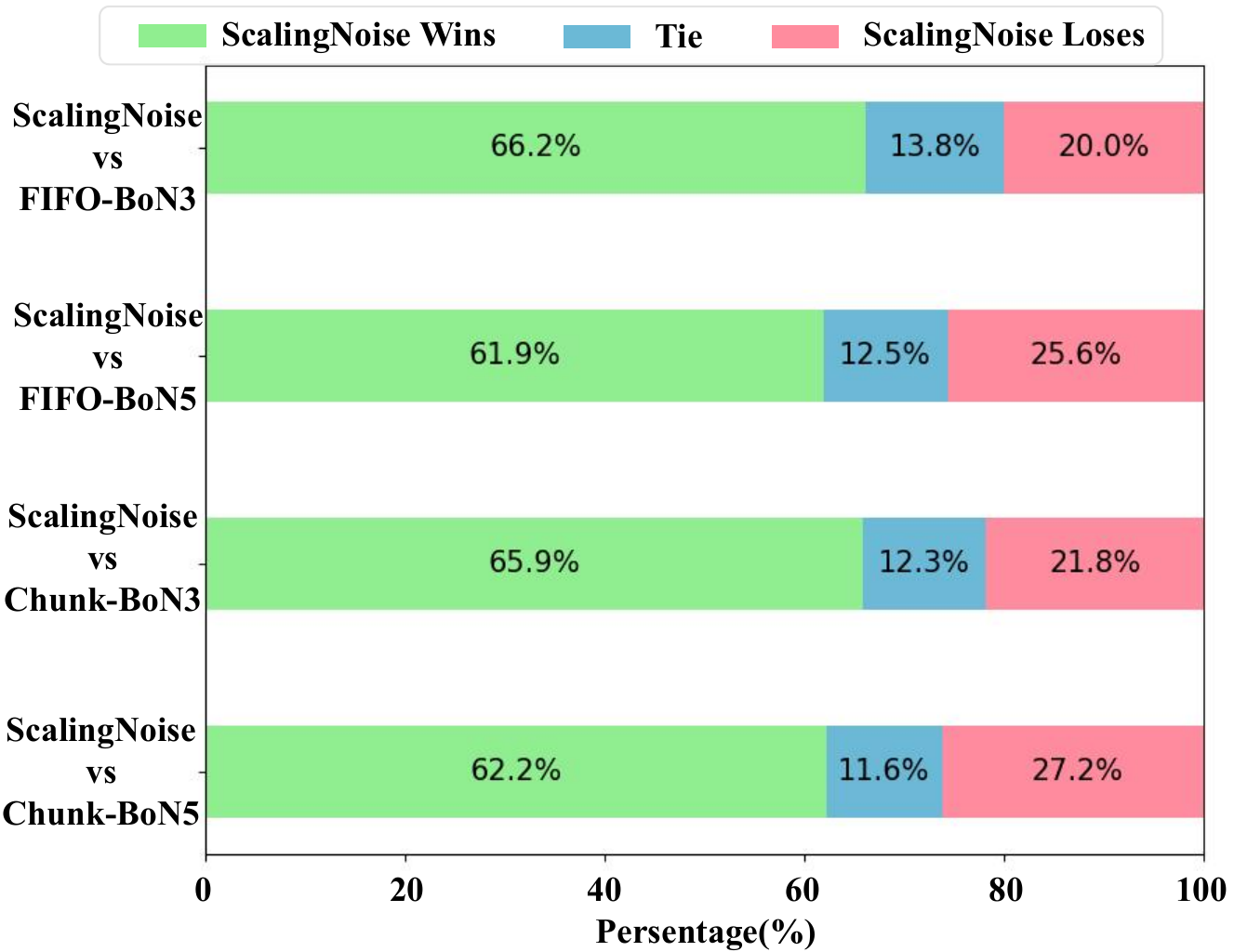}
        \caption{\textbf{User Study.} Win rate of videos generated using \ours compared with other inference-time scaling methods.}
        \label{fig:User Study}
    \end{minipage}
    \hfill % 添加水平间距
    \begin{minipage}{0.48\textwidth}
        \centering
        \includegraphics[width=\textwidth]{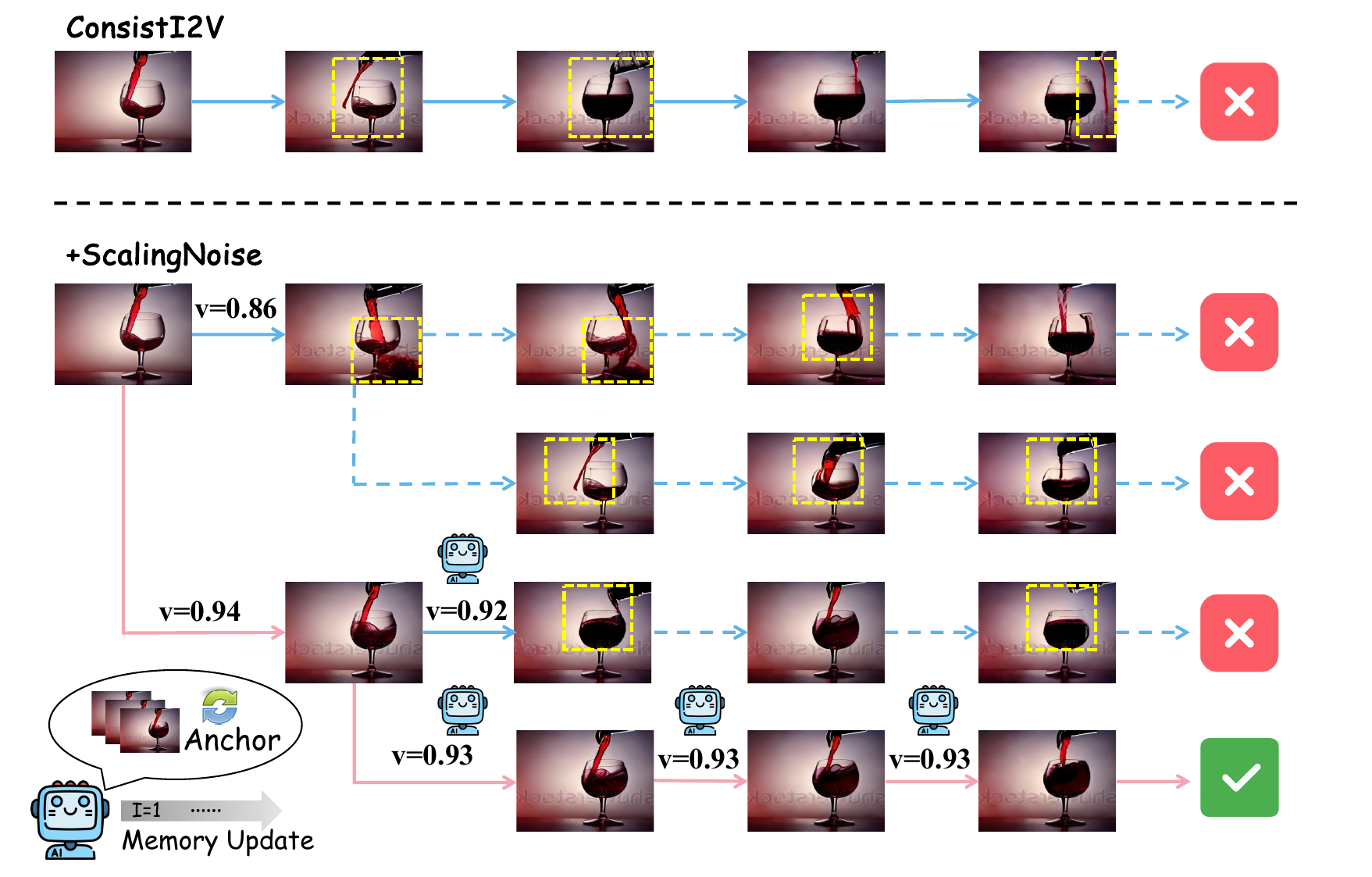}
        \caption{The upper part of this figure represents a greedy approach to generate long videos. In contrast, the tree-structured searching process of \ours is outlined below. Our prompt is “Red wine is poured into a glass. highly detailed, cinematic, arc shot, high contrast, soft lighting".}
        % \caption{The upper part of this figure represents the original method, which involves a greedy approach to generating long videos. In contrast, the tree-structured searching process of \ours is outlined below: We highlight the path from the initial state to the best-performing generation, reporting the score for each node. Our prompt is “Red wine is poured into a glass. highly detailed, cinematic, arc shot, high contrast, soft lighting".}
        \label{fig:case_study}
    \end{minipage}
\vspace{-2mm}
\end{figure}
\begin{table}[!t]
    \centering
    \footnotesize
    \begin{minipage}{0.45\textwidth}
        \vspace{-3mm}
        \centering
        \scalebox{.88}{
        \begin{tabular}{lccc} 
        \toprule
        \textbf{Method}              & \textbf{\makecell{Subjection\\ Consistency$\uparrow$}} & \textbf{\makecell{Overall\\ Score$\uparrow$}} &\textbf{\makecell{Inference\\Time$\downarrow$}}\\
        \midrule
        BoN      & \textbf{97.87}  &   \textbf{92.06} & 477.75    \\
        10-Step  & 97.14   &   91.59 & 79.67   \\
        \rowcolor{gray!30} \ours     & 97.71  &  91.83 & \textbf{12.34}   \\
        \bottomrule
        \end{tabular}}
        \vspace{0.1cm}
        \caption{
        % \small
        \textbf{Reward Function Studies.}
        Video consistency and quality of different reward function guided inference time search.
        }
        \label{tab:effeciency}
    \end{minipage}
    \hfill
    \begin{minipage}{0.45\textwidth}
        \centering
        \hspace{-6.0mm}
        \scalebox{0.88}{
        \begin{tabular}{lccc} 
        \toprule
        \textbf{Method}              & \textbf{\makecell{Subjection\\ Consistency$\uparrow$}} & \textbf{\makecell{Image\\Quality$\uparrow$}}&\textbf{\makecell{Overall\\ Score$\uparrow$}} \\ \midrule
        Local  & 92.16   &   66.83   &   88.94  \\
        Anchor  & 92.67  &   66.39   &   89.28  \\ \rowcolor{gray!30}
        \ours    & \textbf{93.14}   &   \textbf{67.91} &   \textbf{89.60}   \\
        \bottomrule
        \end{tabular}
        }
        \hspace{-5.3mm}
        \vspace{0.1cm}
        \caption{
        % \small
        Video consistency and inference times of different evaluation methods. 
        \ours utilizes one-step evaluation to significantly improve efficiency.
        }
        \label{tab:reward}
    \end{minipage}
    \vspace{-2.3em}
\end{table}

% \begin{wrapfigure}{l}{0.48\textwidth}
%   \centering
%   \includegraphics[width=0.48\textwidth]{figures/user_study.pdf}
%   \caption{\textbf{User Study.} Win rate of videos generated using \ours compared with other inference-time scaling methods.}
%   \label{fig:User Study}
% \end{wrapfigure}
% \begin{wrapfigure}{r}{0.48\textwidth}
%     \centering
%     \includegraphics[width=0.48\textwidth]{figures/case_study.pdf}
%     \caption{The upper part of this figure represents the original method, which involves a greedy approach to generating long videos. In contrast, the tree-structured searching process of \ours is outlined below: We highlight the path from the initial state to the best-performing generation, reporting the score for each node. Our prompt is “Red wine is poured into a glass. highly detailed, cinematic, arc shot, high contrast, soft lighting".}
%     \label{fig:case_study}
% \end{wrapfigure}

\noindent \textbf{One-Step Evaluation's Efficiency and Accuracy.} \label{sec: 4.1.3} To evaluate the computational efficiency and accuracy of our evaluation method, using the VideoCraft2~\cite{chen2024videocrafter2} model, we generate videos with a fixed length, adopting 16 frames. We sample 10 candidate initial noises and employ our one-step evaluation method to select one. We test two baseline approaches: (1) \textbf{BoN}: selection after complete denoising for clarity, (2) \textbf{10-Step Evaluation}: the initial noises are denoised for 10 steps, followed by selection using the same reward model.
As shown in Table~\ref{tab:effeciency}, our one-step evaluation method generates videos in just 12.34 seconds, enabling the assessment and selection of initial noises without compromising baseline performance. The other two baselines require 79.67 and 477.75 seconds, respectively. This efficiency allows for scalable search within the long video generation paradigm.

% \begin{wrapfigure}{r}{0.5\textwidth}
%   \centering
%   \includegraphics[width=0.5\textwidth]{figures/user_study.pdf}
%   \caption{\textbf{User Study.} Win rate of videos generated using \ours compared with other inference-time scaling methods.}
%   \label{fig:User Study}
% \end{wrapfigure}
% \begin{wrapfigure}{r}{0.5\textwidth}
%     \centering
%     \includegraphics[width=0.5\textwidth]{figures/case_study.pdf}
%     \caption{The upper part of this figure represents the original method, which involves a greedy approach to generating long videos. In contrast, the tree-structured searching process of \ours is outlined below: We highlight the path from the initial state to the best-performing generation, reporting the score for each node. Our prompt is “Red wine is poured into a glass. highly detailed, cinematic, arc shot, high contrast, soft lighting".}
%     \label{fig:case_study}
% \end{wrapfigure}
\vspace{-0.3cm}
\subsection{ Qualitative results}
\vspace{-0.3cm}
\noindent \textbf{User Study.}
We start with human evaluation with results shown in Fig.~\ref{fig:User Study}. We utilize generated videos from the evaluation dataset, allowing human annotators to assess and compare the output quality and consistency across different methods. The win rate is then calculated based on their judgments, providing a clear metric for performance comparison. The robust performance of our method, \ours, underscores its capability to produce videos that are not only more natural and visually coherent but also maintain a high level of consistency throughout. Compared to the naive inference time scaling method, BoN, \ours distinctly showcases its superior efficiency. 

\noindent \textbf{Case Study of Search Trajectory.}
As shown in Fig.~\ref{fig:case_study}, \ours demonstrates a clear search process based on consistI2V, which illustrates how it evolves from starting state (contains a prompt and a guided image) into a complete long video. In each step, \ours employs a selection, steering by the long-term reward function. For example, in the first step, the reward model assigns a higher score to images where red wine is not spilled, thus avoiding subsequent cumulative errors. At the same time, it can be seen that due to our long-term reward strategy, even if a bad case has already occurred, our method can still make corrections to subsequent frames based on the anchor frame.

\vspace{-0.3cm}
\subsection{Ablation Study}
\vspace{-0.3cm}
In this section, we conduct ablation studies to evaluate the impact of each design component in \ours for long video generation, including reward models and tilted distribution. Unless otherwise specified, all experiments follow previous settings for a fair comparison.

% \begin{wrapfigure}{l}{0.5\textwidth}
%   \centering
%   \includegraphics[width=0.5\textwidth]{figures/user_study.pdf}
%   \caption{\textbf{User Study.} Win rate of videos generated using \ours compared with other inference-time scaling methods.}
%   \label{fig:User Study}
% \end{wrapfigure}
% \begin{wrapfigure}{r}{0.5\textwidth}
%     \centering
%     \includegraphics[width=0.5\textwidth]{figures/case_study.pdf}
%     \caption{The upper part of this figure represents the original method, which involves a greedy approach to generating long videos. In contrast, the tree-structured searching process of \ours is outlined below: We highlight the path from the initial state to the best-performing generation, reporting the score for each node. Our prompt is “Red wine is poured into a glass. highly detailed, cinematic, arc shot, high contrast, soft lighting".}
%     \label{fig:case_study}
% \end{wrapfigure}

\begin{figure*}[!t]
    \centering
    \includegraphics[width=\linewidth]{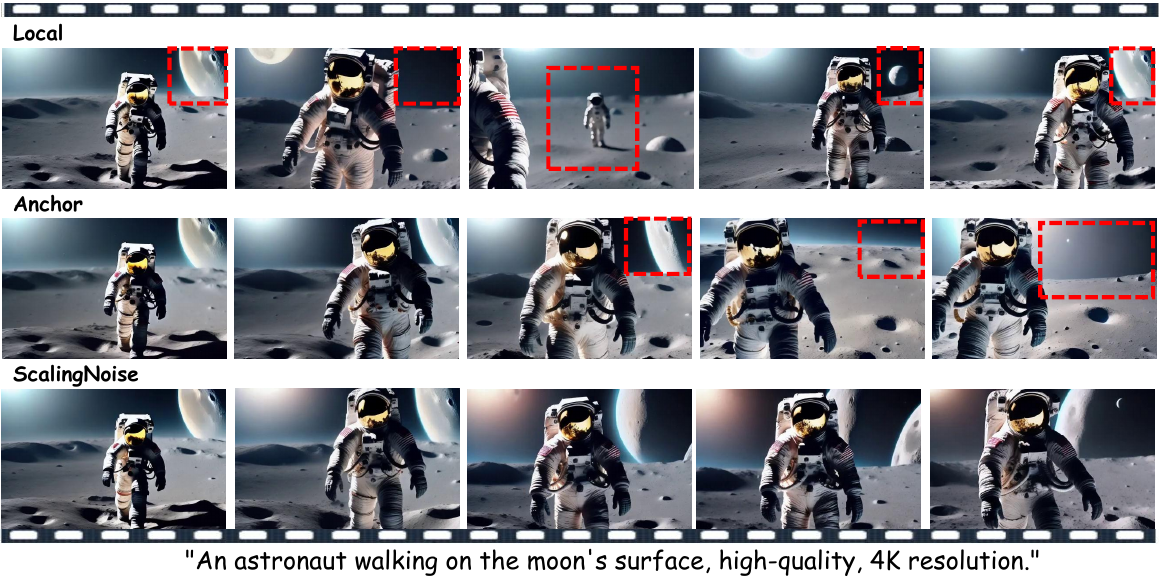}
    \caption{Illustrations of long videos guided by different reward function. \textbf{Row 1}: Local reward only consider the quality of current denoised clip. \textbf{Row 2}: Anchor reward calculate the similarity of the anchor frame and initial noise. \textbf{Row 3}: Ours combines the best of both, achieving long-term reward.}
    \label{reward case}
    \vspace{-2mm}
\end{figure*}

\noindent \textbf{Long-Term Reward.}
First, we present the performance of different variants explored in the design of the reward function based on FIFO. Table~\ref{tab:reward} details the results of two additional runs, while still using the DINO model, with different reward functions: \textit{(1) local reward}: using only local clip during the denoising process, and \textit{(2) anchor reward}: using only the initial noise and anchor frame, leading to a drop of $0.66\%$ and $0.32\%$, respectively.
The specific calculation formula is as follows:
\begin{equation*}
        \Phi_{local} = \sum <d_i\cdot d_{i-1}>, \quad \Phi_{anchor} = <d_a \cdot d_n>.
    \label{equ:reward}
\end{equation*}
As shown in Table~\ref{tab:reward}, \ours achieves the best performance. As illustrated in Fig.~\ref{reward case}, we present videos guided by different reward models. Our reward function not only considers the long-term consistency between the initial noise and anchor frame, but also accounts for the cross-temporal influence of initial noise propagation across video frames within the denoising window.

\begin{figure}[!t]
  \centering
  \includegraphics[width=\linewidth]{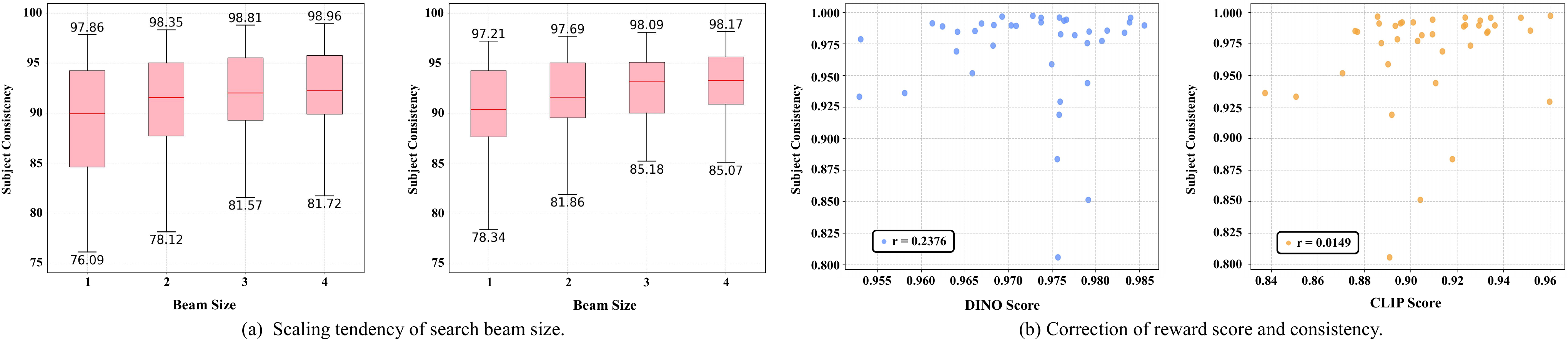}
  \vspace{-0.5cm}
  % \caption{The two images on the left are boxplots showing the effect of increasing scale, with the leftmost based on a FIFO implementation of scale noise and the adjacent one using a chunk-by-chunk approach. The two images on the right depict search results guided by different reward models, with the left one guided by DINO and the right one guided by CLIP.}
  \caption{(a) The two figures are boxplots showing the tendency of scaling beam sizes for ScaleNoise based on two paradigms, in order of FIFO-Diffusion and Chunk by chunk. (b) From Left to Right: Correction of reward model DINO and CLIP feature similarity score and final subject consistency. All points are generated by VideoCraft2.}
    \label{fig:largefig}
  \vspace{-0.2cm}
\end{figure}

% \vspace{3mm}
\noindent \textbf{Different Reward Model.} Then We explored using different reward models(\ie, DINO~\cite{Caron_2021_ICCV} and CLIP~\cite{radford2021clip}) to guide the search process. We generate 16-frame videos and, after one denoising step, scored it using DINO and CLIP. As shown in Fig.~\ref{fig:largefig} (b), the vertical axis represents the subject consistency score, while the horizontal axis represents the reward model scores. It can be observed that DINO’s scores demonstrate a stronger alignment with the final video’s subject consistency compared to CLIP. In contrast to DINO, which effectively captures the features of the primary subject in each frame, CLIP tends to focus on extracting the overall features of the background. During video generation, inconsistencies predominantly stem from variations in the subject, while changes in the background remain relatively minor. Consequently, DINO provides a more accurate and reliable evaluation of subject consistency, making it a superior choice over CLIP for this purpose.

\begin{wraptable}{r}{0.5\textwidth}
% \begin{table}[t]
\vspace{-4mm}
\centering 
\resizebox{0.48\textwidth}{!}{
\begin{tabular}{lccccccc} 
\toprule
\textbf{Method}              & \textbf{\makecell{Subjection\\ Consistency$\uparrow$}} & \textbf{\makecell{Image\\Quality$\uparrow$}}&\textbf{\makecell{Overall\\ Score$\uparrow$}}   \\
\midrule
Random      & 92.64 \textcolor{gray}{$\downarrow$ +0.00}      &   65.98 \textcolor{gray}{$\downarrow$ +0.00} &   89.07 \textcolor{gray}{$\downarrow$ +0.00}  \\
2D FFT          & 92.67 \textcolor{green1}{$\uparrow$ +0.03}   &   65.79 \textcolor{red1}{$\downarrow$ $-$0.19} &  89.24 \textcolor{green1}{$\uparrow$ +0.17} \\
Resample     & 92.94 \textcolor{green1}{$\uparrow$ +0.30}   &   65.71 \textcolor{red1}{$\downarrow$ $-$0.27} &   88.83 \textcolor{red1}{$\downarrow$ $-$0.11}  \\
Reverse     & \textbf{93.27} \textcolor{green1}{$\uparrow$ +0.63}   &   64.83  \textcolor{red1}{$\downarrow$ $-$1.15} &  88.96  \textcolor{red1}{$\downarrow$ $-$0.66} \\ \rowcolor{gray!30}
All (Ours)      & 93.14 \textcolor{green1}{$\uparrow$ +0.50}  &   \textbf{67.91} \textcolor{green1}{$\uparrow$ +1.93} &   \textbf{89.60} \textcolor{green1}{$\uparrow$ +0.53} \\
\bottomrule
\end{tabular}
}
\caption{\textbf{Tilted Distribution studies.} Comparsion of sampling from the  different tilted distribution.}
\vspace{-4mm}
\label{tab:action}
\end{wraptable}
\noindent \textbf{Effectiveness of Tilted Distribution.}
We investigate the tilted distribution impacts on the quality of video generation. Table~\ref{tab:action} summarizes the performance results for long video generation. We tested the performance of these sampling distributions separately, including~\textit{(1) Random Distribution},~\textit{(2) 2D FFT} \textit{(3) DDIM Inversion}~\textit{(4) Inversion Resampling}. The 2D FFT is an effective method for improving video quality. However, as the generated length increases, it can lead to a degradation in video quality. Although the DDIM reverse markedly enhances the  subject consistency, it results in a significant reduction in the range of motion in the generated video. Therefore, we introduce Inversion Resampling to maintain diversity. Integrating all methods into the base model yields a performance boost of +0.53\%.

\vspace{-0.2cm}
\section{Related Work}
\vspace{-0.2cm}
\noindent \textbf{Long viedo generation.}
 Video generation has advanced significantly~\cite{yang2024cogvideox,singer2023makeavideo,blattmann2023stablevideodiffusion,zhou2022magic,uehara2025inferencetimealignmentdiffusionmodels,zheng2024videogenofthoughtcollaborativeframeworkmultishot}, yet producing high-quality long videos remains challenging due to the scarcity of such data and the high computational resources required~\cite{chen2024videocrafter2,blattmann2023stablevideodiffusion,wang2023modelscope}. This limits training models for direct long video generation, leading to the widespread use of autoregressive approaches built on pre-trained models~\cite{deng2025autoregressivevideogenerationvector, xu2023magicanimatetemporallyconsistenthuman}. Current solutions fall into two categories: training-based~\cite{wang2024lingenhighresolutionminutelengthtexttovideo,liu2025luminavideoefficientflexiblevideo,zhang2024triptemporalresiduallearning,si2025repvideorethinkingcrosslayerrepresentation} and training-free methods~\cite{zhao2025riflexfreelunchlength,cai2024ditctrlexploringattentioncontrol,zhang2024chunkwisegenerationlongvideos,yuan2024identitypreservingtexttovideogenerationfrequency}.
Training-based methods~\cite{xiao2025videoauteurlongnarrativevideo,guo2025longcontexttuningvideo} like NUWA-XL~\cite{yin2023nuwa} use a divide-and-conquer strategy to generate long videos layer by layer, while FDM~\cite{harvey2022flexible} and SEINE~\cite{chen2023seine} combine frame interpolation and prediction. Other approaches, such as LDM~\cite{blattmann2023align}, MCVD~\cite{voleti2022mcvd}, Latent-Shift~\cite{an2023latentshiftlatentdiffusiontemporal}, and StreamingT2V~\cite{henschel2024streamingt2v}, incorporate short-term and long-term information as additional inputs. Despite their success, these methods demand high-quality long video data and substantial computation. 
Training-free methods address these challenges. Gen-L-Video~\cite{wang2023genl} and Freenoise~\cite{qiu2024freenoise} use a chunk-by-chunk approach, linking segments with final frames, but this risks degradation and inconsistency. Freelong~\cite{lu2024freelong} blends global and local data via high-low frequency decoupling, while FreeInit~\cite{wu2025freeinit} refines initial noises for better consistency. FIFO-Diffusion~\cite{kim2025fifo} introduces a novel paradigm, reorganizing denoising with a noise-level queue, dequeuing clear frames, enqueuing noise, for efficient, flexible long video generation. In this work, we propose a plug-and-play inference-time strategy that can improve the consistency of videos based on these long video generation methods.

\noindent \textbf{Inference-time Search.}
A variety of inference-time search strategies have been proven crucial in long context generation within the field of LLMs~\cite{yoon2025montecarlotreediffusion,huang2024o1replicationjourney,li2025llmseasilylearnreason,tangintervening,xue2025mmrc,liu2024unveilingignorancemllmsseeing}. 
The advent of DeepSeek-R1~\cite{deepseekai2025deepseekr1incentivizingreasoningcapability} has further advanced inference-time search. 
By applying various search techniques in the language space, such as controlled decoding~\cite{chakraborty2024transfer, xu2024genarm, guo2024coldattackjailbreakingllmsstealthiness,yu2025flowreasoningtrainingllmsdivergent}, best of N~\cite{lightman2023letsverifystepstep, li2024common7blanguagemodels}, and Monte Carlo tree search~\cite{zhang2024rest, tian2024toward, wang2024litesearch, wang2024towards}, LLMs achieve better step-level responses, thus enhancing performance.
During inference-time search, leveraging a good process reward model (PRM)~\cite{uesato2022solving,cobbe2021training,hosseini2024vstartrainingverifiersselftaught,wang2024mathshepherdverifyreinforcellms} is essential to determine the quality of the responses.
\cite{ma2025inferencetimescalingdiffusionmodels} proposed using supervised verifiers as a signal to guide generating trajectories within Diffusion Models (DMs), but did not investigate its impact on video generation inference-time search. Furthermore, some work has preliminarily explored inference-time search in VDMs~\cite{uehara2025inferencetimealignmentdiffusionmodels,xie2025sana15efficientscaling,zekri2025finetuningdiscretediffusionmodels,tian2025diffusionsharpeningfinetuningdiffusionmodels}, however, there is still a lack of investigation into the inference-time scaling law and long-term signals in the process of long video generation. In this work, we explore the effectiveness of scaling inference-time  budget utilizing beam search to enhance the consistency of generated long videos.

\vspace{-0.3cm}
\section{Conclusion \& Limitation}
\vspace{-0.3cm}
In this work, we introduce \ours, a novel inference-time search strategy that significantly enhances the consistency of VDMs by identifying golden initial noises to optimize video generation. Utilizing a guided one-step denoising process and a reward model anchored to prior content, \ours achieves superior global content coherence while maintaining high-level object features across multi-chunk video sequences. Furthermore, by integrating a tilted noise distribution, it facilitates more effective exploration of the state space, further elevating generation quality. Our findings show that scaling inference-time computations enhances both video consistency and the quality of individual frames. Experiments on benchmarks validate that ScalingNoise substantially enhances content fidelity and subject consistency in resource-constrained long video generation.

% \vspace{3mm}
\noindent\textbf{Limitation:} We clarify the limitations of our proposed \ours: \textit{(i):} Our \ours may struggle with scenes involving highly complex or abrupt motion, where accurate alignment across frames becomes challenging, potentially affecting temporal coherence. \textit{(ii)}: \ours cannot completely eliminate accumulated error, while through long-term signal guidance, it can to some extent alleviate this phenomenon. To entirely address this issue, we need to conduct an extra in-depth analysis of the causes of error accumulation.

{
    \small
    \bibliographystyle{plainnat}
    \bibliography{main}
}

%%%%%%%%%%%%%%%%%%%%%%%%%%%%%%%%%%%%%%%%%%%%%%%%%%%%%%%%%%%%

\newpage
\section*{NeurIPS Paper Checklist}

\begin{enumerate}

\item {\bf Claims}
    \item[] Question: Do the main claims made in the abstract and introduction accurately reflect the paper's contributions and scope?
    \item[] Answer: \answerYes{} % Replace by \answerYes{}, \answerNo{}, or \answerNA{}.
    \item[] Justification: We claim our contributions and scope in Introduction.
    \item[] Guidelines:
    \begin{itemize}
        \item The answer NA means that the abstract and introduction do not include the claims made in the paper.
        \item The abstract and/or introduction should clearly state the claims made, including the contributions made in the paper and important assumptions and limitations. A No or NA answer to this question will not be perceived well by the reviewers. 
        \item The claims made should match theoretical and experimental results, and reflect how much the results can be expected to generalize to other settings. 
        \item It is fine to include aspirational goals as motivation as long as it is clear that these goals are not attained by the paper. 
    \end{itemize}

\item {\bf Limitations}
    \item[] Question: Does the paper discuss the limitations of the work performed by the authors?
    \item[] Answer: \answerYes{} % Replace by \answerYes{}, \answerNo{}, or \answerNA{}.
    \item[] Justification: We discuss the limitations of our limitations in the last section, Conclusion\&Limitation.
    \item[] Guidelines:
    \begin{itemize}
        \item The answer NA means that the paper has no limitation while the answer No means that the paper has limitations, but those are not discussed in the paper. 
        \item The authors are encouraged to create a separate "Limitations" section in their paper.
        \item The paper should point out any strong assumptions and how robust the results are to violations of these assumptions (e.g., independence assumptions, noiseless settings, model well-specification, asymptotic approximations only holding locally). The authors should reflect on how these assumptions might be violated in practice and what the implications would be.
        \item The authors should reflect on the scope of the claims made, e.g., if the approach was only tested on a few datasets or with a few runs. In general, empirical results often depend on implicit assumptions, which should be articulated.
        \item The authors should reflect on the factors that influence the performance of the approach. For example, a facial recognition algorithm may perform poorly when image resolution is low or images are taken in low lighting. Or a speech-to-text system might not be used reliably to provide closed captions for online lectures because it fails to handle technical jargon.
        \item The authors should discuss the computational efficiency of the proposed algorithms and how they scale with dataset size.
        \item If applicable, the authors should discuss possible limitations of their approach to address problems of privacy and fairness.
        \item While the authors might fear that complete honesty about limitations might be used by reviewers as grounds for rejection, a worse outcome might be that reviewers discover limitations that aren't acknowledged in the paper. The authors should use their best judgment and recognize that individual actions in favor of transparency play an important role in developing norms that preserve the integrity of the community. Reviewers will be specifically instructed to not penalize honesty concerning limitations.
    \end{itemize}

\item {\bf Theory assumptions and proofs}
    \item[] Question: For each theoretical result, does the paper provide the full set of assumptions and a complete (and correct) proof?
    \item[] Answer: \answerNA{} % Replace by \answerYes{}, \answerNo{}, or \answerNA{}.
    \item[] Justification: \answerNA{}
    \item[] Guidelines:
    \begin{itemize}
        \item The answer NA means that the paper does not include theoretical results. 
        \item All the theorems, formulas, and proofs in the paper should be numbered and cross-referenced.
        \item All assumptions should be clearly stated or referenced in the statement of any theorems.
        \item The proofs can either appear in the main paper or the supplemental material, but if they appear in the supplemental material, the authors are encouraged to provide a short proof sketch to provide intuition. 
        \item Inversely, any informal proof provided in the core of the paper should be complemented by formal proofs provided in appendix or supplemental material.
        \item Theorems and Lemmas that the proof relies upon should be properly referenced. 
    \end{itemize}

    \item {\bf Experimental result reproducibility}
    \item[] Question: Does the paper fully disclose all the information needed to reproduce the main experimental results of the paper to the extent that it affects the main claims and/or conclusions of the paper (regardless of whether the code and data are provided or not)?
    \item[] Answer: \answerYes{} % Replace by \answerYes{}, \answerNo{}, or \answerNA{}.
    \item[] Justification: We clearly provide the information of our experiment in the section, Baseline and Implementation details. 
    \item[] Guidelines:
    \begin{itemize}
        \item The answer NA means that the paper does not include experiments.
        \item If the paper includes experiments, a No answer to this question will not be perceived well by the reviewers: Making the paper reproducible is important, regardless of whether the code and data are provided or not.
        \item If the contribution is a dataset and/or model, the authors should describe the steps taken to make their results reproducible or verifiable. 
        \item Depending on the contribution, reproducibility can be accomplished in various ways. For example, if the contribution is a novel architecture, describing the architecture fully might suffice, or if the contribution is a specific model and empirical evaluation, it may be necessary to either make it possible for others to replicate the model with the same dataset, or provide access to the model. In general. releasing code and data is often one good way to accomplish this, but reproducibility can also be provided via detailed instructions for how to replicate the results, access to a hosted model (e.g., in the case of a large language model), releasing of a model checkpoint, or other means that are appropriate to the research performed.
        \item While NeurIPS does not require releasing code, the conference does require all submissions to provide some reasonable avenue for reproducibility, which may depend on the nature of the contribution. For example
        \begin{enumerate}
            \item If the contribution is primarily a new algorithm, the paper should make it clear how to reproduce that algorithm.
            \item If the contribution is primarily a new model architecture, the paper should describe the architecture clearly and fully.
            \item If the contribution is a new model (e.g., a large language model), then there should either be a way to access this model for reproducing the results or a way to reproduce the model (e.g., with an open-source dataset or instructions for how to construct the dataset).
            \item We recognize that reproducibility may be tricky in some cases, in which case authors are welcome to describe the particular way they provide for reproducibility. In the case of closed-source models, it may be that access to the model is limited in some way (e.g., to registered users), but it should be possible for other researchers to have some path to reproducing or verifying the results.
        \end{enumerate}
    \end{itemize}

\item {\bf Open access to data and code}
    \item[] Question: Does the paper provide open access to the data and code, with sufficient instructions to faithfully reproduce the main experimental results, as described in supplemental material?
    \item[] Answer: \answerYes{} % Replace by \answerYes{}, \answerNo{}, or \answerNA{}.
    \item[] Justification: We provide open access to the code to reproduce wo method.
    \item[] Guidelines:
    \begin{itemize}
        \item The answer NA means that paper does not include experiments requiring code.
        \item Please see the NeurIPS code and data submission guidelines (\url{https://nips.cc/public/guides/CodeSubmissionPolicy}) for more details.
        \item While we encourage the release of code and data, we understand that this might not be possible, so “No” is an acceptable answer. Papers cannot be rejected simply for not including code, unless this is central to the contribution (e.g., for a new open-source benchmark).
        \item The instructions should contain the exact command and environment needed to run to reproduce the results. See the NeurIPS code and data submission guidelines (\url{https://nips.cc/public/guides/CodeSubmissionPolicy}) for more details.
        \item The authors should provide instructions on data access and preparation, including how to access the raw data, preprocessed data, intermediate data, and generated data, etc.
        \item The authors should provide scripts to reproduce all experimental results for the new proposed method and baselines. If only a subset of experiments are reproducible, they should state which ones are omitted from the script and why.
        \item At submission time, to preserve anonymity, the authors should release anonymized versions (if applicable).
        \item Providing as much information as possible in supplemental material (appended to the paper) is recommended, but including URLs to data and code is permitted.
    \end{itemize}

\item {\bf Experimental setting/details}
    \item[] Question: Does the paper specify all the training and test details (e.g., data splits, hyperparameters, how they were chosen, type of optimizer, etc.) necessary to understand the results?
    \item[] Answer: \answerYes{} % Replace by \answerYes{}, \answerNo{}, or \answerNA{}.
    \item[] Justification: We clearly provide our experiment setting in the section, Baseline and Implementation details.
    \item[] Guidelines:
    \begin{itemize}
        \item The answer NA means that the paper does not include experiments.
        \item The experimental setting should be presented in the core of the paper to a level of detail that is necessary to appreciate the results and make sense of them.
        \item The full details can be provided either with the code, in appendix, or as supplemental material.
    \end{itemize}

\item {\bf Experiment statistical significance}
    \item[] Question: Does the paper report error bars suitably and correctly defined or other appropriate information about the statistical significance of the experiments?
    \item[] Answer: \answerNA{} % Replace by \answerYes{}, \answerNo{}, or \answerNA{}.
    \item[] Justification: \answerNA{}
    \item[] Guidelines:
    \begin{itemize}
        \item The answer NA means that the paper does not include experiments.
        \item The authors should answer "Yes" if the results are accompanied by error bars, confidence intervals, or statistical significance tests, at least for the experiments that support the main claims of the paper.
        \item The factors of variability that the error bars are capturing should be clearly stated (for example, train/test split, initialization, random drawing of some parameter, or overall run with given experimental conditions).
        \item The method for calculating the error bars should be explained (closed form formula, call to a library function, bootstrap, etc.)
        \item The assumptions made should be given (e.g., Normally distributed errors).
        \item It should be clear whether the error bar is the standard deviation or the standard error of the mean.
        \item It is OK to report 1-sigma error bars, but one should state it. The authors should preferably report a 2-sigma error bar than state that they have a 96\% CI, if the hypothesis of Normality of errors is not verified.
        \item For asymmetric distributions, the authors should be careful not to show in tables or figures symmetric error bars that would yield results that are out of range (e.g. negative error rates).
        \item If error bars are reported in tables or plots, The authors should explain in the text how they were calculated and reference the corresponding figures or tables in the text.
    \end{itemize}

\item {\bf Experiments compute resources}
    \item[] Question: For each experiment, does the paper provide sufficient information on the computer resources (type of compute workers, memory, time of execution) needed to reproduce the experiments?
    \item[] Answer: \answerYes{} % Replace by \answerYes{}, \answerNo{}, or \answerNA{}.
    \item[] Justification: We provide sufficient information on the computation resources in Baseline and Implementation details.
    \item[] Guidelines:
    \begin{itemize}
        \item The answer NA means that the paper does not include experiments.
        \item The paper should indicate the type of compute workers CPU or GPU, internal cluster, or cloud provider, including relevant memory and storage.
        \item The paper should provide the amount of compute required for each of the individual experimental runs as well as estimate the total compute. 
        \item The paper should disclose whether the full research project required more compute than the experiments reported in the paper (e.g., preliminary or failed experiments that didn't make it into the paper). 
    \end{itemize}
    
\item {\bf Code of ethics}
    \item[] Question: Does the research conducted in the paper conform, in every respect, with the NeurIPS Code of Ethics \url{https://neurips.cc/public/EthicsGuidelines}?
    \item[] Answer: \answerYes{} % Replace by \answerYes{}, \answerNo{}, or \answerNA{}.
    \item[] Justification: This paper strictly adheres to all requirements of the NeurIPS Code of Ethics, including transparency in data usage, fairness in research methods, with relevant details provided in Section 3.2.
    \item[] Guidelines:
    \begin{itemize}
        \item The answer NA means that the authors have not reviewed the NeurIPS Code of Ethics.
        \item If the authors answer No, they should explain the special circumstances that require a deviation from the Code of Ethics.
        \item The authors should make sure to preserve anonymity (e.g., if there is a special consideration due to laws or regulations in their jurisdiction).
    \end{itemize}

\item {\bf Broader impacts}
    \item[] Question: Does the paper discuss both potential positive societal impacts and negative societal impacts of the work performed?
    \item[] Answer: \answerNA{} % Replace by \answerYes{}, \answerNo{}, or \answerNA{}.
    \item[] Justification: \answerNA{}
    \item[] Guidelines:
    \begin{itemize}
        \item The answer NA means that there is no societal impact of the work performed.
        \item If the authors answer NA or No, they should explain why their work has no societal impact or why the paper does not address societal impact.
        \item Examples of negative societal impacts include potential malicious or unintended uses (e.g., disinformation, generating fake profiles, surveillance), fairness considerations (e.g., deployment of technologies that could make decisions that unfairly impact specific groups), privacy considerations, and security considerations.
        \item The conference expects that many papers will be foundational research and not tied to particular applications, let alone deployments. However, if there is a direct path to any negative applications, the authors should point it out. For example, it is legitimate to point out that an improvement in the quality of generative models could be used to generate deepfakes for disinformation. On the other hand, it is not needed to point out that a generic algorithm for optimizing neural networks could enable people to train models that generate Deepfakes faster.
        \item The authors should consider possible harms that could arise when the technology is being used as intended and functioning correctly, harms that could arise when the technology is being used as intended but gives incorrect results, and harms following from (intentional or unintentional) misuse of the technology.
        \item If there are negative societal impacts, the authors could also discuss possible mitigation strategies (e.g., gated release of models, providing defenses in addition to attacks, mechanisms for monitoring misuse, mechanisms to monitor how a system learns from feedback over time, improving the efficiency and accessibility of ML).
    \end{itemize}
    
\item {\bf Safeguards}
    \item[] Question: Does the paper describe safeguards that have been put in place for responsible release of data or models that have a high risk for misuse (e.g., pretrained language models, image generators, or scraped datasets)?
    \item[] Answer: \answerNA{} % Replace by \answerYes{}, \answerNo{}, or \answerNA{}.
    \item[] Justification: \answerNA{}
    \item[] Guidelines:
    \begin{itemize}
        \item The answer NA means that the paper poses no such risks.
        \item Released models that have a high risk for misuse or dual-use should be released with necessary safeguards to allow for controlled use of the model, for example by requiring that users adhere to usage guidelines or restrictions to access the model or implementing safety filters. 
        \item Datasets that have been scraped from the Internet could pose safety risks. The authors should describe how they avoided releasing unsafe images.
        \item We recognize that providing effective safeguards is challenging, and many papers do not require this, but we encourage authors to take this into account and make a best faith effort.
    \end{itemize}

\item {\bf Licenses for existing assets}
    \item[] Question: Are the creators or original owners of assets (e.g., code, data, models), used in the paper, properly credited and are the license and terms of use explicitly mentioned and properly respected?
    \item[] Answer: \answerYes{} % Replace by \answerYes{}, \answerNo{}, or \answerNA{}.
    \item[] Justification: We explicitly acknowledge the original owners of the assets, including code, data, and models.
    \item[] Guidelines:
    \begin{itemize}
        \item The answer NA means that the paper does not use existing assets.
        \item The authors should cite the original paper that produced the code package or dataset.
        \item The authors should state which version of the asset is used and, if possible, include a URL.
        \item The name of the license (e.g., CC-BY 4.0) should be included for each asset.
        \item For scraped data from a particular source (e.g., website), the copyright and terms of service of that source should be provided.
        \item If assets are released, the license, copyright information, and terms of use in the package should be provided. For popular datasets, \url{paperswithcode.com/datasets} has curated licenses for some datasets. Their licensing guide can help determine the license of a dataset.
        \item For existing datasets that are re-packaged, both the original license and the license of the derived asset (if it has changed) should be provided.
        \item If this information is not available online, the authors are encouraged to reach out to the asset's creators.
    \end{itemize}

\item {\bf New assets}
    \item[] Question: Are new assets introduced in the paper well documented and is the documentation provided alongside the assets?
    \item[] Answer: \answerNA{} % Replace by \answerYes{}, \answerNo{}, or \answerNA{}.
    \item[] Justification: \answerNA{}
    \item[] Guidelines:
    \begin{itemize}
        \item The answer NA means that the paper does not release new assets.
        \item Researchers should communicate the details of the dataset/code/model as part of their submissions via structured templates. This includes details about training, license, limitations, etc. 
        \item The paper should discuss whether and how consent was obtained from people whose asset is used.
        \item At submission time, remember to anonymize your assets (if applicable). You can either create an anonymized URL or include an anonymized zip file.
    \end{itemize}

\item {\bf Crowdsourcing and research with human subjects}
    \item[] Question: For crowdsourcing experiments and research with human subjects, does the paper include the full text of instructions given to participants and screenshots, if applicable, as well as details about compensation (if any)? 
    \item[] Answer: \answerNA{} % Replace by \answerYes{}, \answerNo{}, or \answerNA{}.
    \item[] Justification: \answerNA{}
    \item[] Guidelines:
    \begin{itemize}
        \item The answer NA means that the paper does not involve crowdsourcing nor research with human subjects.
        \item Including this information in the supplemental material is fine, but if the main contribution of the paper involves human subjects, then as much detail as possible should be included in the main paper. 
        \item According to the NeurIPS Code of Ethics, workers involved in data collection, curation, or other labor should be paid at least the minimum wage in the country of the data collector. 
    \end{itemize}

\item {\bf Institutional review board (IRB) approvals or equivalent for research with human subjects}
    \item[] Question: Does the paper describe potential risks incurred by study participants, whether such risks were disclosed to the subjects, and whether Institutional Review Board (IRB) approvals (or an equivalent approval/review based on the requirements of your country or institution) were obtained?
    \item[] Answer: \answerNA{} % Replace by \answerYes{}, \answerNo{}, or \answerNA{}.
    \item[] Justification: \answerNA{}
    \item[] Guidelines:
    \begin{itemize}
        \item The answer NA means that the paper does not involve crowdsourcing nor research with human subjects.
        \item Depending on the country in which research is conducted, IRB approval (or equivalent) may be required for any human subjects research. If you obtained IRB approval, you should clearly state this in the paper. 
        \item We recognize that the procedures for this may vary significantly between institutions and locations, and we expect authors to adhere to the NeurIPS Code of Ethics and the guidelines for their institution. 
        \item For initial submissions, do not include any information that would break anonymity (if applicable), such as the institution conducting the review.
    \end{itemize}

\item {\bf Declaration of LLM usage}
    \item[] Question: Does the paper describe the usage of LLMs if it is an important, original, or non-standard component of the core methods in this research? Note that if the LLM is used only for writing, editing, or formatting purposes and does not impact the core methodology, scientific rigorousness, or originality of the research, declaration is not required.
    %this research? 
    \item[] Answer: \answerYes{} % Replace by \answerYes{}, \answerNo{}, or \answerNA{}.
    \item[] Justification: We describe the details of how our model is used as the vision encoder, along with the corresponding experimental settings.
    \item[] Guidelines:
    \begin{itemize}
        \item The answer NA means that the core method development in this research does not involve LLMs as any important, original, or non-standard components.
        \item Please refer to our LLM policy (\url{https://neurips.cc/Conferences/2025/LLM}) for what should or should not be described.
    \end{itemize}

\end{enumerate}

\appendix
\newpage

\section{Algorithm of ScalingNoise}
\label{appendix:A}
This section illustrates pseudo-code for \ours.
\begin{algorithm}[h]
\caption{\ours Inference-time Search}
\label{alg: ScalingNoise}
\begin{algorithmic}[1]
\REQUIRE{
Diffusion Model $D$, Reward Function $\Phi$, Sample Tilted Distribution $Sample$, Condition $\cond$, Beam Size $k$, Step Size $n$, DDIM Steps $\tau_t$, Generated Video $\bm{V}=[\ ]$} Anchor Frame $\bm{v}_a$
\WHILE{Generation is not Done}
\FOR{i in [1, 2, ..., k]}
    \STATE $\bm{r} = [\ ]$
    \FOR{j in [1, 2, ..., n]}
        \STATE{$\bm{\epsilon}_{ij} \gets Sample(\bm{V}$)}
        \STATE{$\hat{\bm{v}}_{ij} \gets D(\bm{\epsilon}_{ij}, \cond,\text{num\_steps}=\tau_t)$}
        \STATE{$r_{ij} \gets \Phi(\hat{\bm{v}}_{ij}, \bm{v}_a)$}
        \STATE{$\bm{r}$.append($r_{ij}$)}
        \ENDFOR
    \ENDFOR
\STATE{$[\bm{v}_1, \dots, \bm{v}_k] \gets \text{Select the best } k \text{ elements from } \bm{r}$}
\FOR{i in [$\tau_0$, $\tau_1$, ..., $\tau_t$]}
    \STATE{$\bm{v} \gets D(\bm{\epsilon}, \cond, \text{num\_steps}=i)$}
    \ENDFOR
\STATE {$\bm{v}_a \gets \bm{v}_{i0}$}
\STATE Append current clip $[\bm{v}_1, \dots, \bm{v}_k]$ to $\bm{V}$
\ENDWHILE
\RETURN{$\bm{V}$}
\end{algorithmic}
\end{algorithm}
% \vspace{-0.5cm}

\section{Baseline}
% \textbf{Baslines.}
Our approach is benchmarked against several methods:
\begin{itemize}[leftmargin=0.5cm]
% \item \textbf{Naïve latent extend}
% The most naive method, which involves extending the initial noise and utilizing the original pipeline to generate videos longer than those in the training set.

\item \textbf{FreeNoise}~\cite{qiu2024freenoise}:
We chose FreeNoise as a baseline because it is also a training-free method that can base the VideoCrafter2~\cite{chen2024videocrafter2} model, which also serves as our base model, to generate long videos. 
It employs a rescheduling technique for the initialization noise and incorporates Window-based Attention Fusion to generate longer videos.

\item \textbf{Streaming T2V}~\cite{henschel2024streamingt2v}:
To assess our method's effectiveness in generating longer videos, StreamingT2V was chosen as our baseline. 
Streaming T2V involves training a new model that uses an auto-regressive approach to produce long-form videos. Our prompt is "Red wine is poured into a glass. highly detailed, cinematic, arc shot, high contrast, soft lighting, 4k resolution.
A spectacular fireworks display over Sydney Harbour, 4K, high resolution.".

% \item \textbf{OpenSora V1.1}~\cite{hpcaitech2024opensora}: As a video diffusion model based on DiT~\cite{peebles2023scalable}, OpenSora V1.1 supports up to 120 frames and is capable of generating videos at various resolutions. 
\item \textbf{OpenSora V1.1}~\cite{hpcaitech2024opensora}: a video diffusion model based on DiT~\cite{peebles2023scalable}, supports up to 120 frames, can generate videos at various resolutions, and has been specifically trained on longer video sequences to enhance its extended video generation capabilities.
\end{itemize}
% \section{}
\section{Benchmark}
\noindent \textbf{Vbench.} Following is the detail of the five evaluation metrics in our paper:
Subject Consistency assesses the uniformity and coherence of the primary subject across frames using DINO~\cite{Caron_2021_ICCV} features. Background Consistency is measured by the CLIP~\cite{radford2021learning} feature similarity. Temporal Flickering~\cite{teed2020raft} evaluates the frame-wise consistency and Motion Smoothness~\cite{li2023amt} assesses the fluidity and jittering of motion. Finally, we use MUSIQ~\cite{ke2021musiq} to predict the image quality which mainly considers the low-level distortions presented in the generated video frames.

\newpage
\section{VideoCrafter2}
\label{app:qual:vc2}
In Fig.~\ref{fig:qual:vc1_0} and Fig.~\ref{fig:qual:vc2_0}, we provide more qualitative results with VideoCrafter2~\cite{chen2024videocrafter2}.\\

\scalebox{1}{
    \setlength{\tabcolsep}{1pt}
    \hspace{-5mm}
    \begin{tabular}{ccccc} \\
        \includegraphics[width=0.2\linewidth]{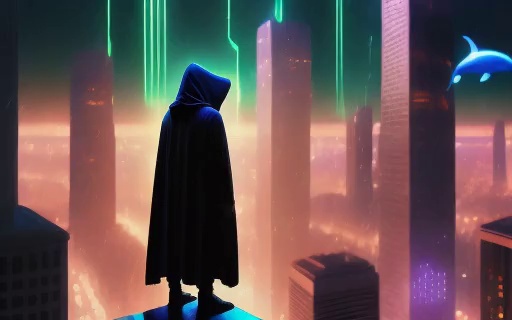} &
        \includegraphics[width=0.2\linewidth]{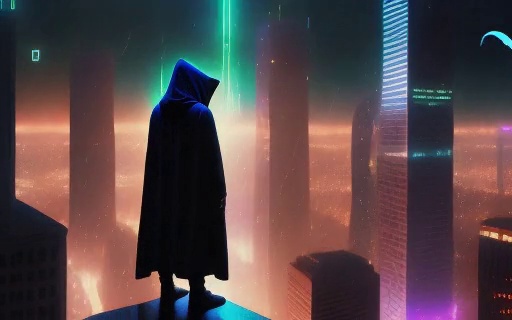} &
        \includegraphics[width=0.2\linewidth]{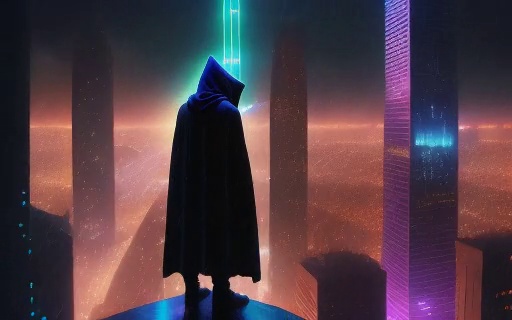} &
        \includegraphics[width=0.2\linewidth]{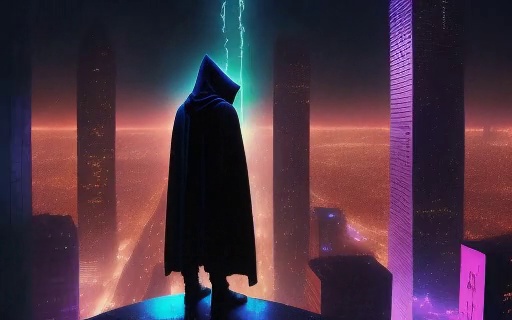} &
        \includegraphics[width=0.2\linewidth]{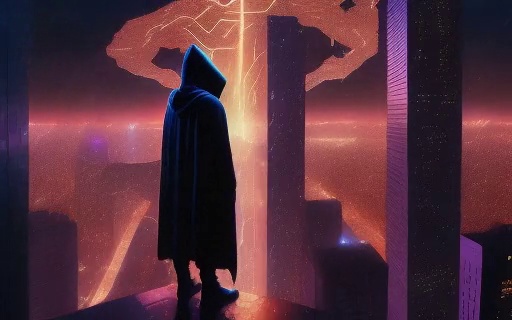} \\
        \multicolumn{5}{c}{\small \parbox{0.8\textwidth} {\center  (a) \textsf{"A lone figure in a hooded cloak stands atop a skyscraper, city lights sprawling below like glowing circuit board."}}} \vspace{2pt} \\
        \includegraphics[width=0.2\linewidth]{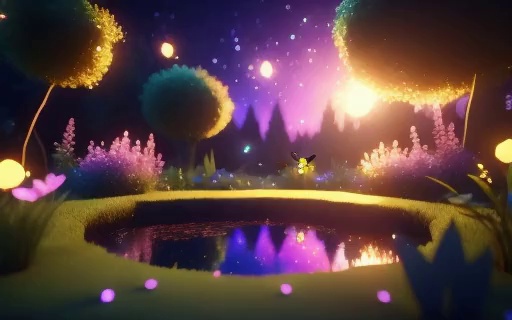} &
        \includegraphics[width=0.2\linewidth]{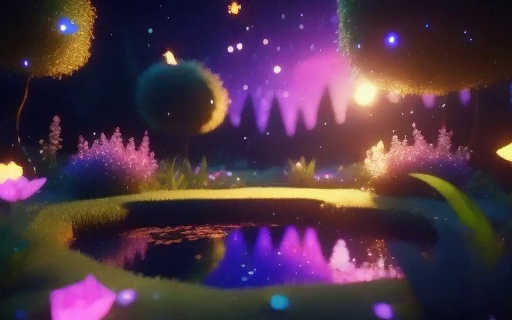} &
        \includegraphics[width=0.2\linewidth]{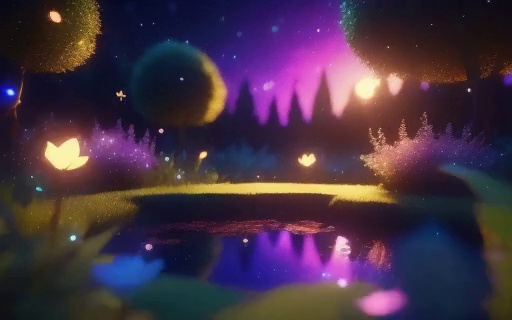} &
        \includegraphics[width=0.2\linewidth]{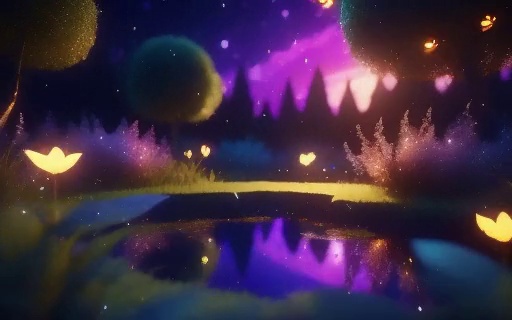} &
        \includegraphics[width=0.2\linewidth]{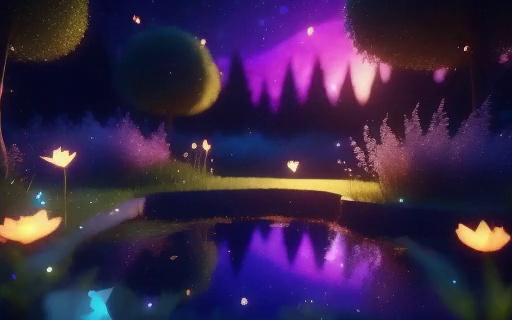} \\
        \multicolumn{5}{c}{\small \parbox{0.8\textwidth} {\center  (b) \textsf{"A mystical, low-poly enchanted garden with glowing plants and a small pond \\where fireflies dance under a starry sky."}}}\vspace{2pt} \\
        \includegraphics[width=0.2\linewidth]{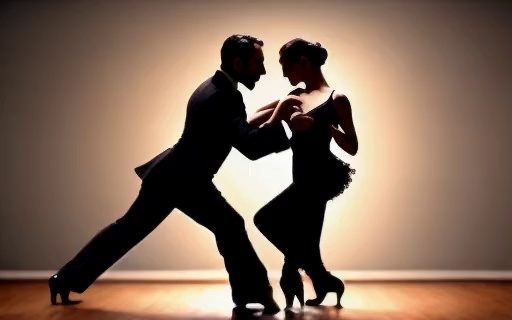} &
        \includegraphics[width=0.2\linewidth]{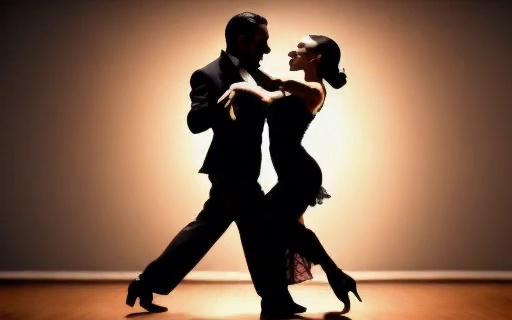} &
        \includegraphics[width=0.2\linewidth]{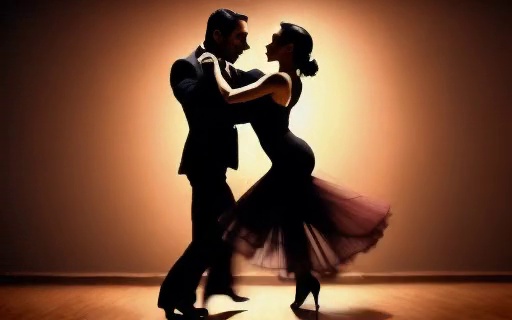} &
        \includegraphics[width=0.2\linewidth]{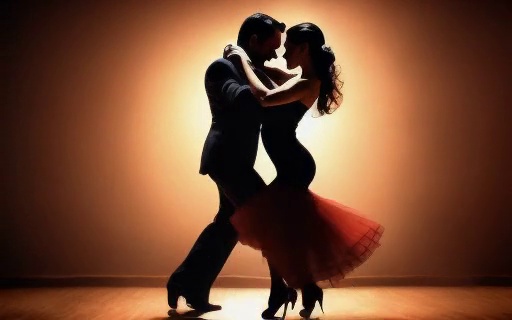} &
        \includegraphics[width=0.2\linewidth]{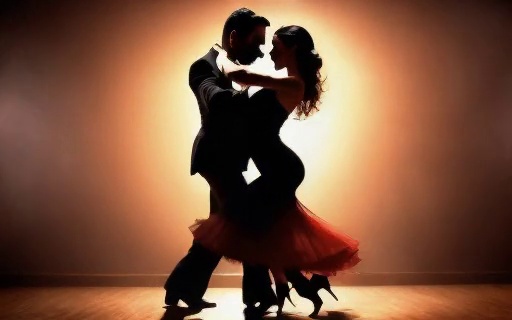} \\
        \multicolumn{5}{c}{\small \parbox{0.8\textwidth} {\center  (c) \textsf{"A pair of tango dancers performing in Buenos Aires, 4K, high resolution."}}}\vspace{2pt} \\
        \includegraphics[width=0.2\linewidth]{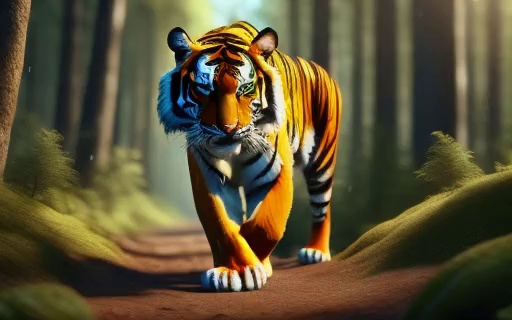} &
        \includegraphics[width=0.2\linewidth]{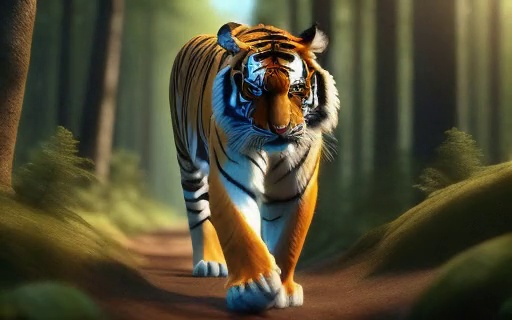} &
        \includegraphics[width=0.2\linewidth]{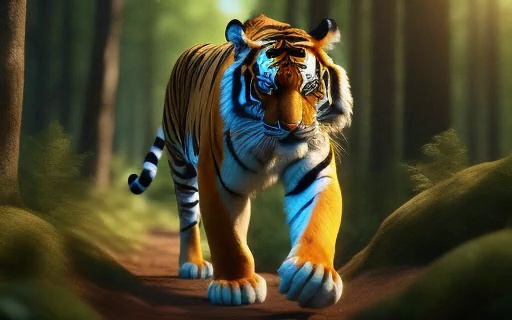} &
        \includegraphics[width=0.2\linewidth]{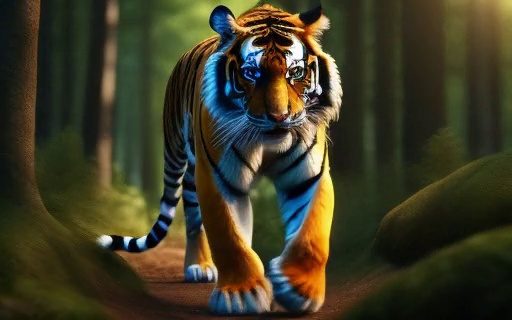} &
        \includegraphics[width=0.2\linewidth]{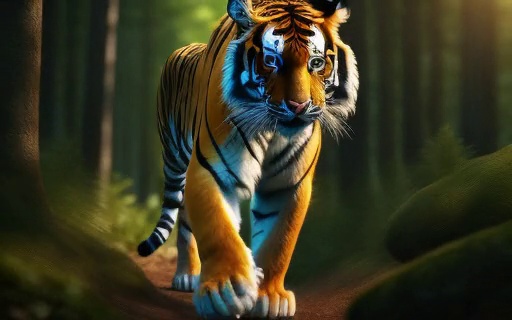} \\
        \multicolumn{5}{c}{\small (d) \textsf{"A tiger walks in the forest, photorealistic, 4k, high definition, 4k resolution."}}\vspace{2pt} \\
        \includegraphics[width=0.2\linewidth]{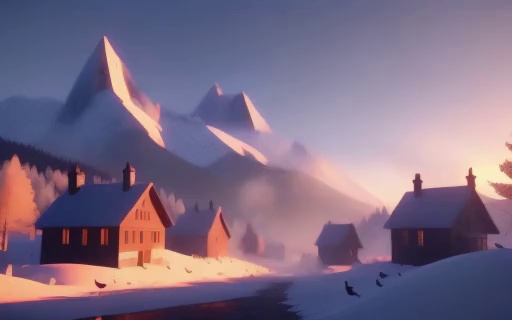} &
        \includegraphics[width=0.2\linewidth]{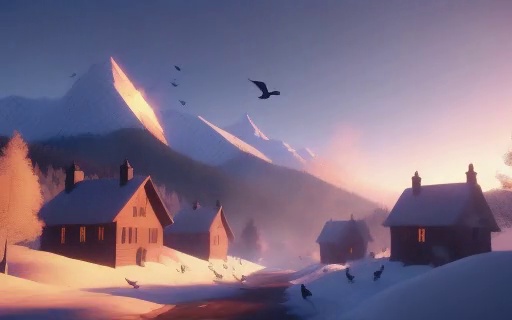} &
        \includegraphics[width=0.2\linewidth]{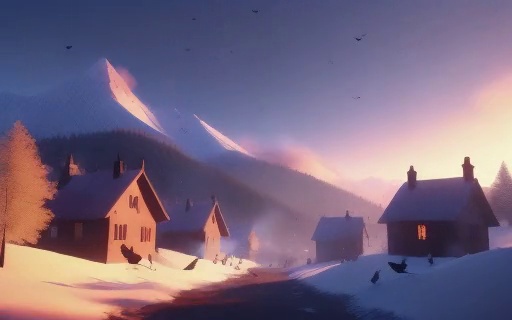} &
        \includegraphics[width=0.2\linewidth]{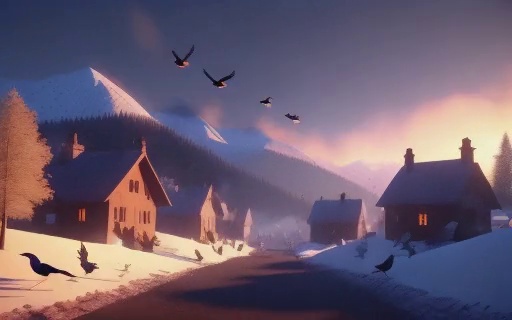} &
        \includegraphics[width=0.2\linewidth]{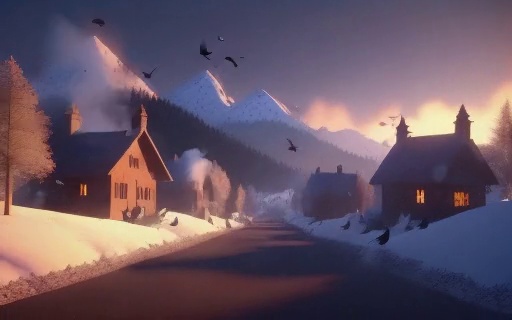} \\
        \multicolumn{5}{c}{\small \parbox{0.8\textwidth} {\center (e) \textsf{"A tranquil, low-poly mountain village at dawn, \\with smoke rising from chimneys and birds flying over snowy peaks."}}}\vspace{2pt} \\

        \includegraphics[width=0.2\linewidth]{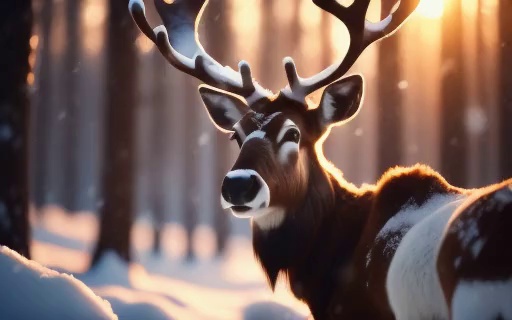} &
        \includegraphics[width=0.2\linewidth]{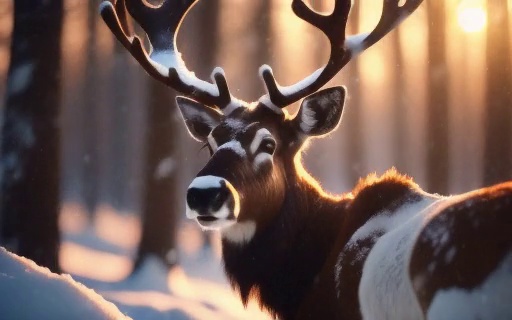} &
        \includegraphics[width=0.2\linewidth]{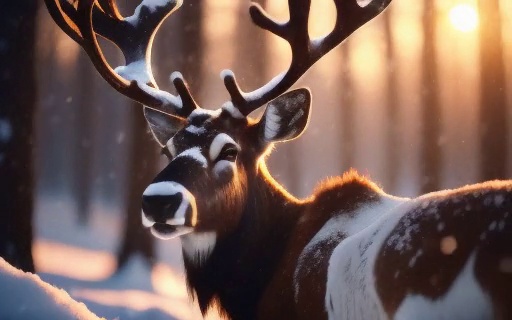} &
        \includegraphics[width=0.2\linewidth]{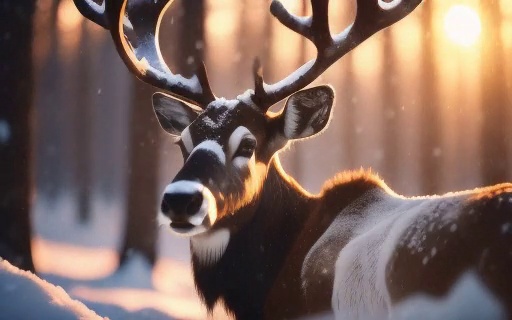} &
        \includegraphics[width=0.2\linewidth]{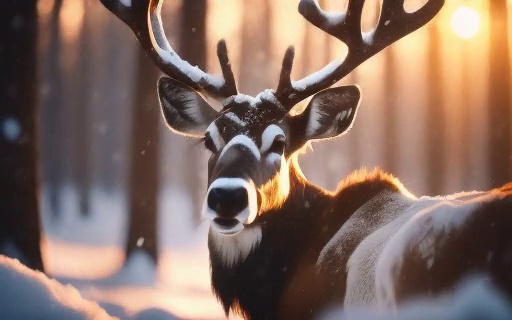} \\
        \multicolumn{5}{c}{\small (f) \textsf{"Cinematic closeup and detailed portrait of a reindeer in a snowy forest at sunset."}}\vspace{2pt}
    \end{tabular}
}
\captionof{figure}{
    Videos generated by \ours with VideoCrafter2 based on the paradigm of FIFO-Diffusion.
    }
\label{fig:qual:vc1_0}

\newpage
\scalebox{1}{
    \setlength{\tabcolsep}{1pt}
    \hspace{-5mm}
    \begin{tabular}{ccccc} \\
        \includegraphics[width=0.2\linewidth]{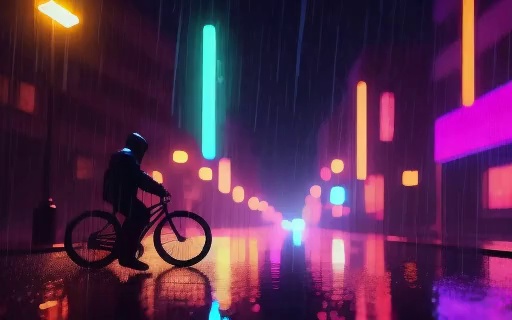} &
        \includegraphics[width=0.2\linewidth]{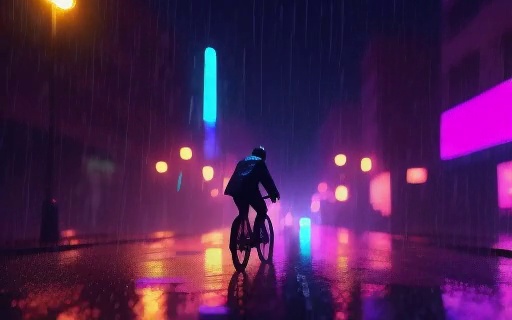} &
        \includegraphics[width=0.2\linewidth]{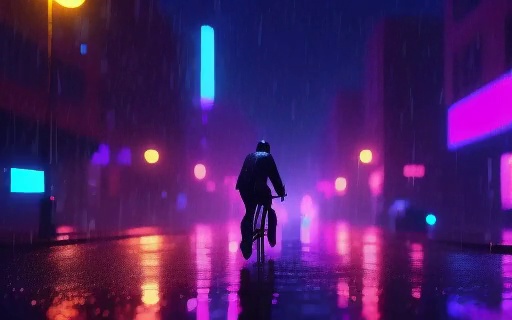} &
        \includegraphics[width=0.2\linewidth]{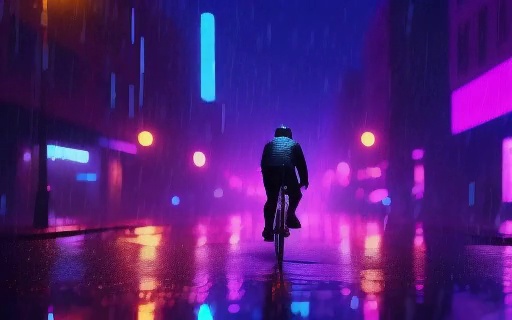} &
        \includegraphics[width=0.2\linewidth]{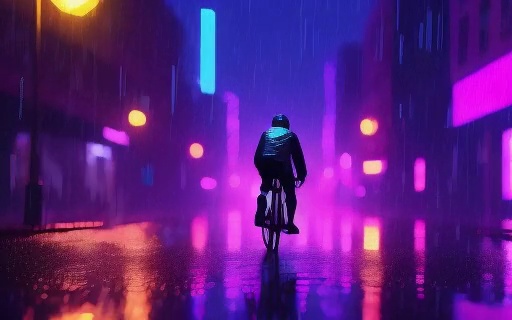} \\
        \multicolumn{5}{c}{\small \parbox{0.8\textwidth} {\center  (a) \textsf{"A dynamic, low-poly cityscape at night, \\with neon lights reflecting off wet streets and a lone cyclist riding through the rain."}}} \vspace{2pt} \\
        \includegraphics[width=0.2\linewidth]{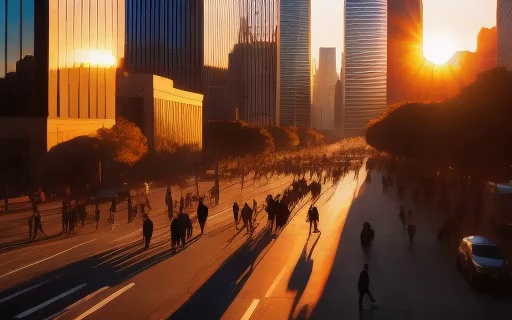} &
        \includegraphics[width=0.2\linewidth]{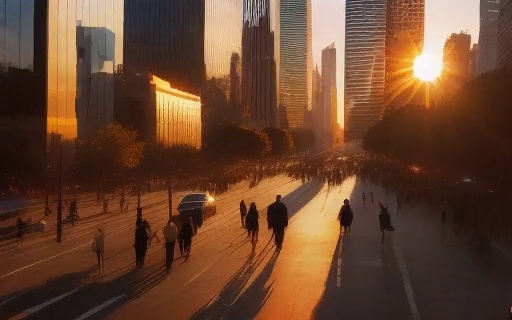} &
        \includegraphics[width=0.2\linewidth]{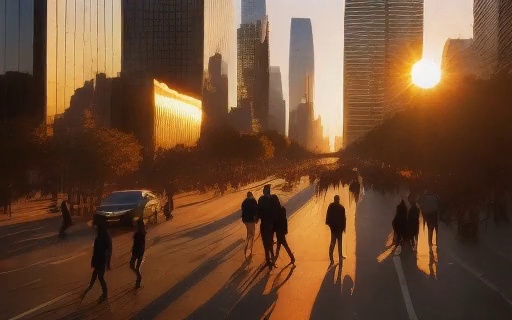} &
        \includegraphics[width=0.2\linewidth]{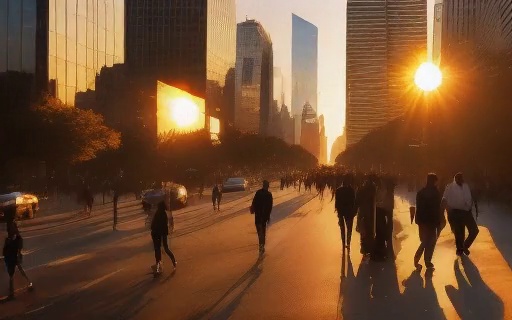} &
        \includegraphics[width=0.2\linewidth]{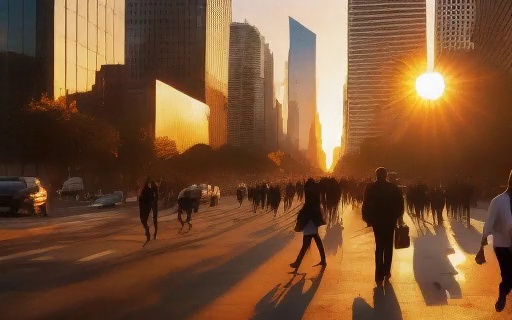} \\
        \multicolumn{5}{c}{\small \parbox{0.8\textwidth} {\center  (b) \textsf{"A bustling cityscape at sunset with skyscrapers reflecting golden light, \\people walking, and traffic moving swiftly."}}}\vspace{2pt} \\
        \includegraphics[width=0.2\linewidth]{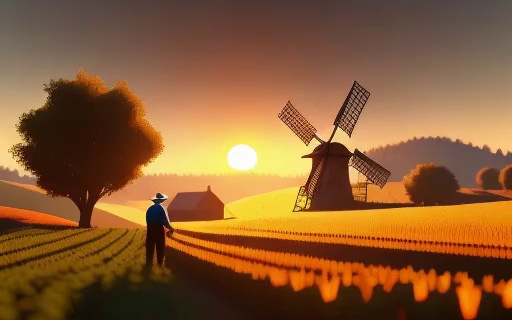} &
        \includegraphics[width=0.2\linewidth]{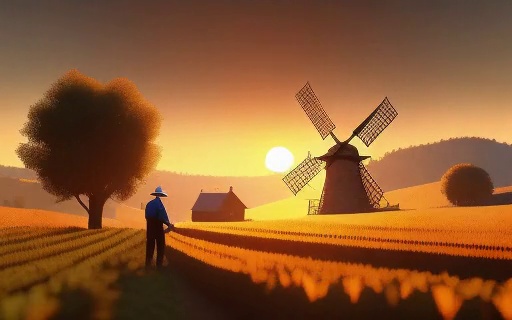} &
        \includegraphics[width=0.2\linewidth]{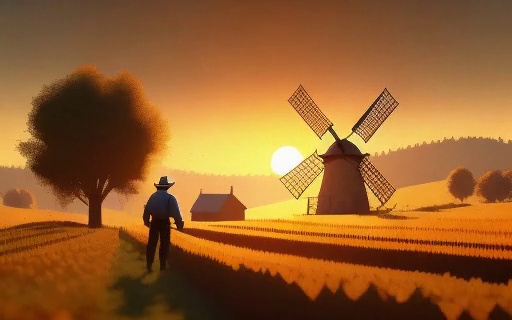} &
        \includegraphics[width=0.2\linewidth]{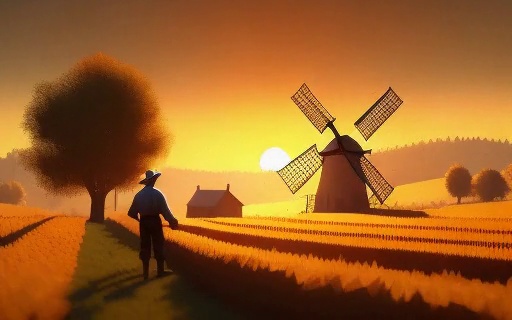} &
        \includegraphics[width=0.2\linewidth]{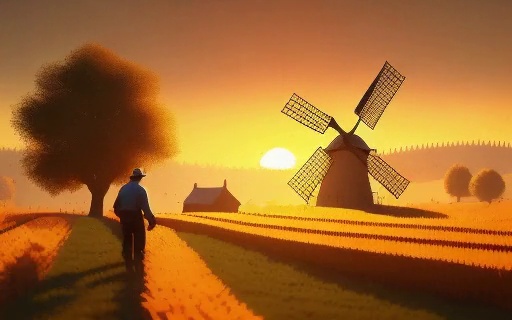} \\
        \multicolumn{5}{c}{\small \parbox{0.8\textwidth} {\center  (c) \textsf{"A peaceful, low-poly countryside with rolling hills, a windmill, \\and a farmer tending to his crops under a golden sunset."}}}\vspace{2pt} \\
        \includegraphics[width=0.2\linewidth]{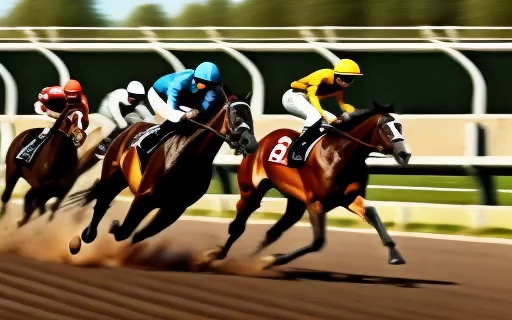} &
        \includegraphics[width=0.2\linewidth]{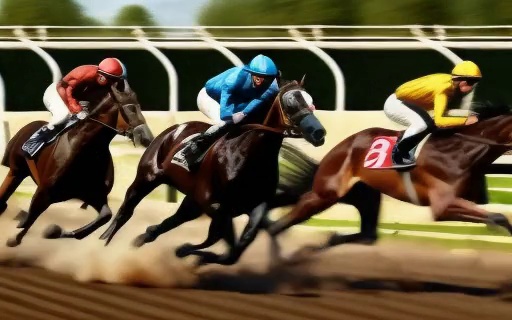} &
        \includegraphics[width=0.2\linewidth]{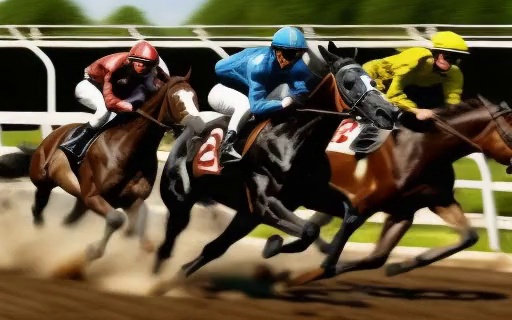} &
        \includegraphics[width=0.2\linewidth]{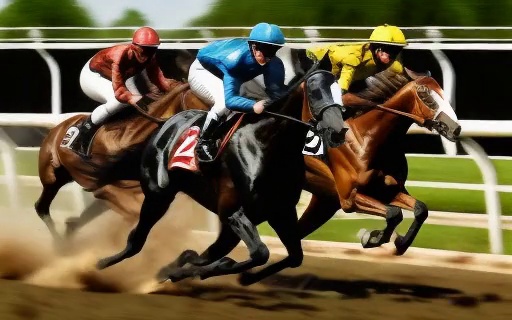} &
        \includegraphics[width=0.2\linewidth]{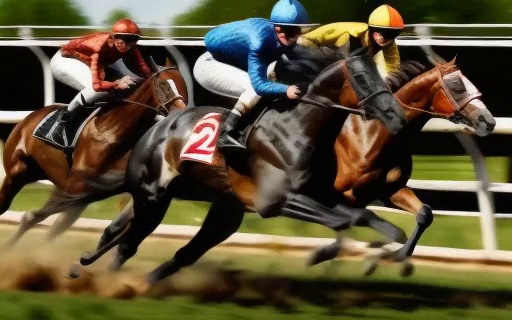} \\
        \multicolumn{5}{c}{\small (d) \textsf{"A horse race in full gallop, capturing the speed and excitement, 2K, photorealistic."}}\vspace{2pt} \\
        \includegraphics[width=0.2\linewidth]{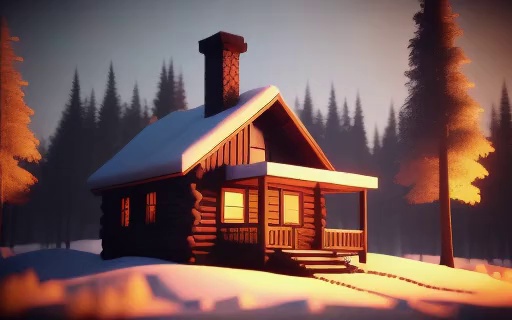} &
        \includegraphics[width=0.2\linewidth]{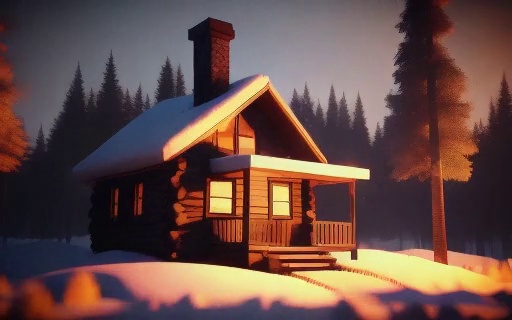} &
        \includegraphics[width=0.2\linewidth]{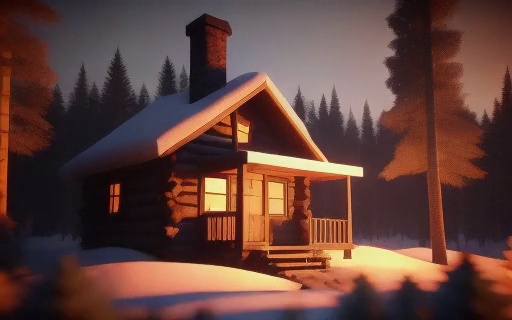} &
        \includegraphics[width=0.2\linewidth]{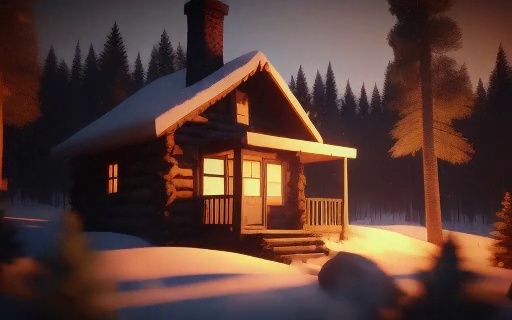} &
        \includegraphics[width=0.2\linewidth]{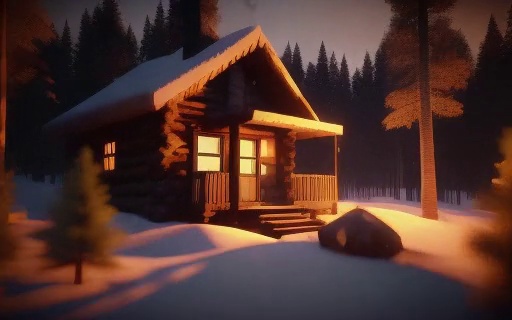} \\
        \multicolumn{5}{c}{\small \parbox{0.9\textwidth}{\centering (e) \textsf{"A cozy, low-poly cabin in the woods surrounded by tall pine trees, with a warm light \\glowing from the windows and smoke curling from the chimney, 4k resolution."}}}\vspace{2pt} \\
        
        \includegraphics[width=0.2\linewidth]{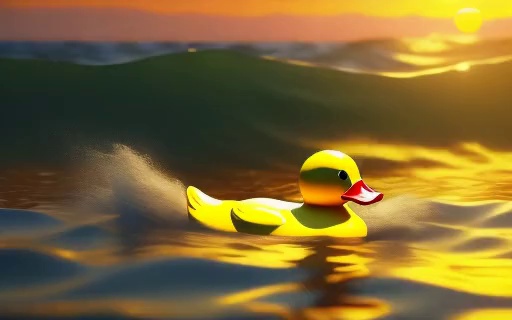} &
        \includegraphics[width=0.2\linewidth]{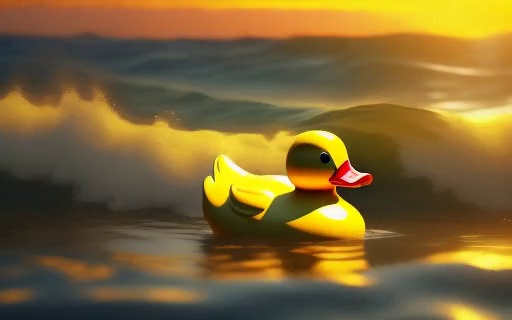} &
        \includegraphics[width=0.2\linewidth]{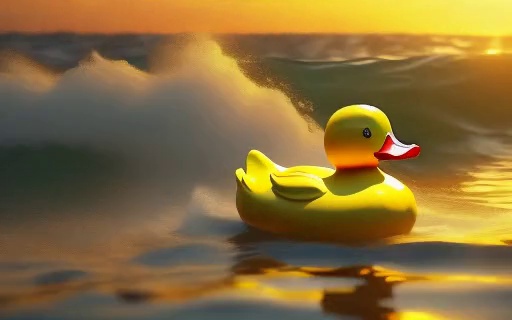} &
        \includegraphics[width=0.2\linewidth]{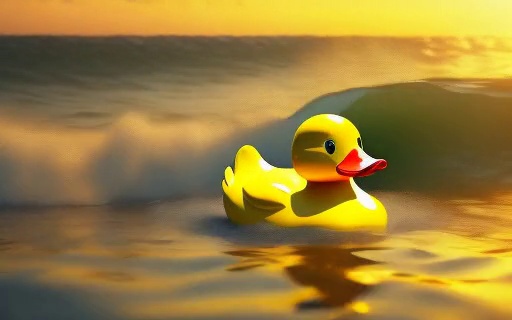} &
        \includegraphics[width=0.2\linewidth]{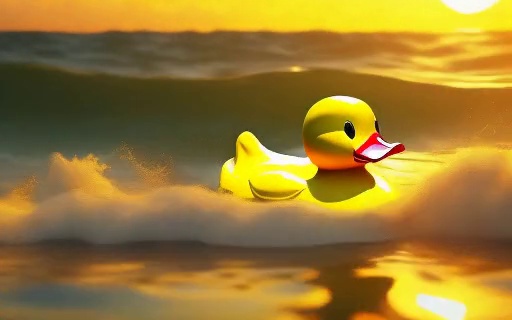} \\
        \multicolumn{5}{c}{\small (f) \textsf{"Impressionist style, a yellow rubber duck floating on the wave on the sunset, 4k resolution."}}\vspace{2pt}
    \end{tabular}
}
\captionof{figure}{
    Videos generated by \ours with VideoCrafter2 based on the paradigm of FIFO-Diffusion.
    }
\label{fig:qual:vc2_0}

\begin{figure}[h]
    \centering 
    \includegraphics[width=1\textwidth]{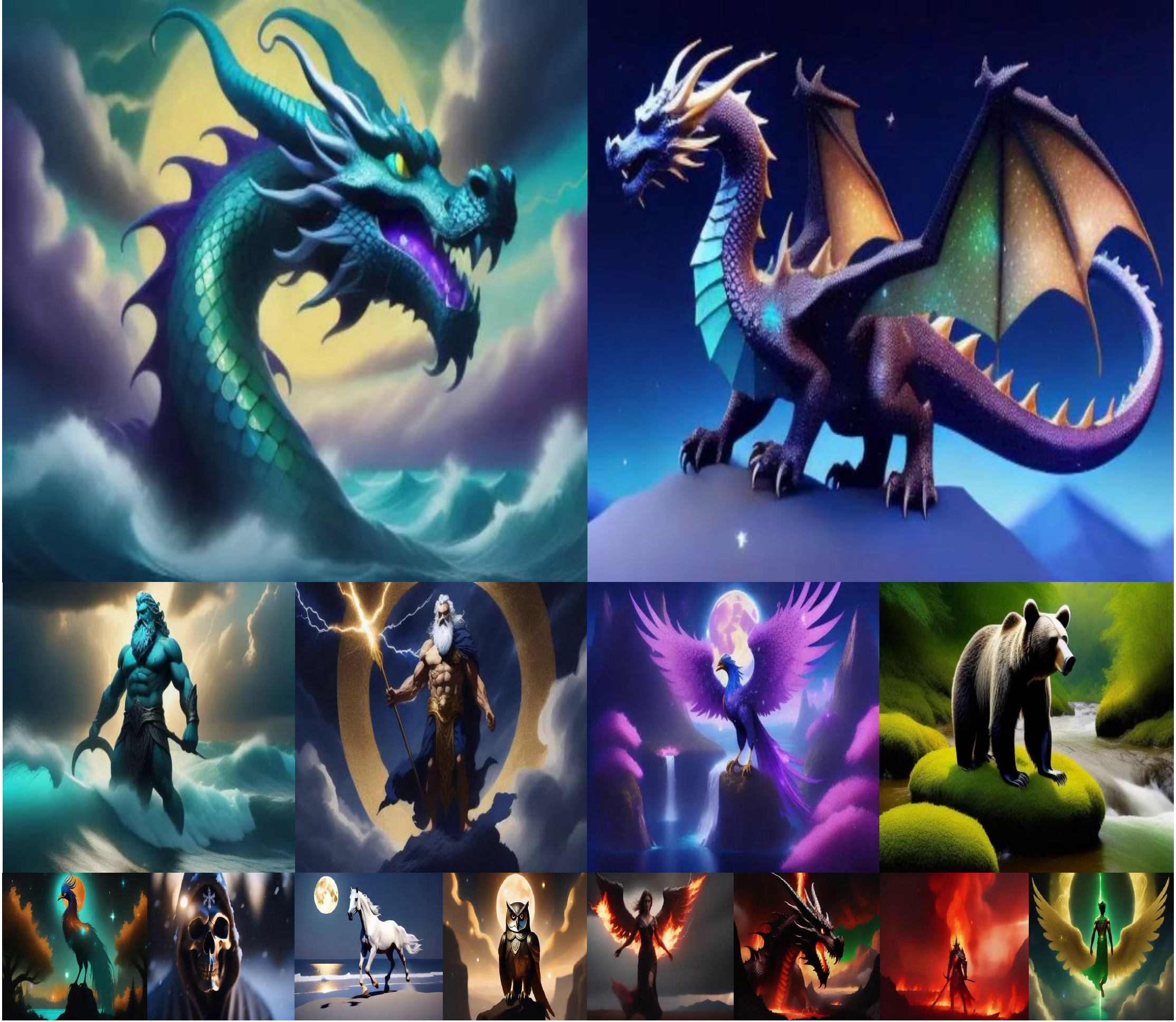} 
    \caption{The last individual video frame generated by \ours with VideoCrafter2 based on the paradigm of FIFO-Diffusion.} 
    \label{fig:img_case}
\end{figure}

\newpage
In Fig.~\ref{fig:img_case}, we present the last individual frame of the generated videos. \\

\end{document}